\documentclass{article}
\usepackage{arxiv}
\usepackage{newtxmath}
\usepackage[utf8]{inputenc} 
\usepackage[T1]{fontenc}    
\usepackage{hyperref}       
\usepackage{url}            
\usepackage{booktabs}       
\usepackage{amsfonts}       
\usepackage{nicefrac}       
\usepackage{microtype}      
\usepackage{lipsum}		
\usepackage{graphicx}
\usepackage{natbib}
\usepackage{doi}

\title{An explainable three dimensional framework to uncover learning patterns: A unified look in variable sulci recognition}
\date{July 9, 2024}	
\author{Michail ~Mamalakis\\
Corresponding author: mm2703@cam.ac.uk\\
Department of Psychiatry \\ 
Department of Computer Science and Technology\\
University of Cambridge, Cambridge, UK.\\
\And
Héloïse ~de Vareilles\\
Department of Psychiatry \\ 
University of Cambridge, Cambridge, UK.\\
\And
Atheer  ~Al-Manea\\
Department of Psychiatry \\ 
University of Cambridge, Cambridge, UK.\\
\And
Samantha C. ~Mitchell\\
Department of Psychology \\ 
University of Cambridge, Cambridge, UK.\\
\And
Ingrid ~Agartz\\
Department of Psychiatric Research\\
Diakonhjemmet Hospital, Oslo, Norway\\
\And
Lynn Egeland ~Mørch-Johnsen\\
Department of Psychiatry \\
Department of Clinical Research\\
Østfold Hospital, Grålum, Norway.\\
\And
Jane  ~Garrison\\
Department of Psychology \\ 
University of Cambridge, Cambridge, UK.\\
\And
Jon ~Simons\\
Department of Psychology \\ 
University of Cambridge, Cambridge, UK.\\
\And
Pietro ~Lio\\
Department of Computer Science and Technology\\
University of Cambridge, Cambridge, UK.\\
\And
John ~Suckling\\
Department of Psychiatry \\ 
University of Cambridge, Cambridge, UK.\\
\And
Graham ~Murray\\
Department of Psychiatry \\ 
University of Cambridge, Cambridge, UK.\\
}

\renewcommand{\shorttitle}{An explainable three dimensional framework to uncover learning patterns.}

\begin{document}
\maketitle

\begin{abstract}
The significant features identified in a representative subset of the dataset during the learning process of an artificial intelligence model are referred to as a 'global' explanation. Three-dimensional (3D) global explanations are crucial in neuroimaging, where a complex representational space demands more than basic two-dimensional interpretations. However, current studies in the literature often lack the accuracy, comprehensibility, and 3D global explanations needed in neuroimaging and beyond.
To address this gap, we developed an explainable artificial intelligence (XAI) 3D-Framework capable of providing accurate, low-complexity global explanations. We evaluated the framework using various 3D deep learning models trained on a well-annotated cohort of 596 structural MRIs. The binary classification task focused on detecting the presence or absence of the paracingulate sulcus (PCS), a highly variable brain structure associated with psychosis.
Our framework integrates statistical features (Shape) and XAI methods (GradCam and SHAP) with dimensionality reduction, ensuring that explanations reflect both model learning and cohort-specific variability. By combining Shape, GradCam, and SHAP, our framework reduces inter-method variability, enhancing the faithfulness and reliability of global explanations. These robust explanations facilitated the identification of critical sub-regions, including the posterior temporal and internal parietal regions, as well as the cingulate region and thalamus, suggesting potential genetic or developmental influences.
For the first time, this XAI 3D-Framework leverages global explanations to uncover the broader developmental context of specific cortical features. This approach advances the fields of deep learning and neuroscience by offering insights into normative brain development and atypical trajectories linked to mental illness, paving the way for more reliable and interpretable AI applications in neuroimaging.
\end{abstract}


\section{Introduction}
In both medical imaging and neuroscience, explainability holds paramount importance. Recently, the study by \cite{perspective} introduced the necessity of explanations in artificial intelligence (AI) healthcare applications, categorizing them into four types: self-explainable, semi-explainable, non-explainable applications, and new-pattern discovery. This categorization is based on the variability of expert opinions, the stability of the evaluation protocol, and the dimensionality of the problem.
\par In neuroscience AI applications, there commonly exists significant variability in evaluation protocols and decision-making among experts. The inherent high uncertainty creates a greater need for explainability (as the framework falls in the non-explainable category; \cite{perspective}). To this end, applications usually need both "local" methods, which provide explanations for each AI prediction separately, and "global" methods, which derive explanations for the decision-making of the AI across the entire dataset. Such variability underscores the importance of thorough investigation and validation of AI model behavior. Even among experienced professionals, knowledge gaps can persist, and this is where AI has the potential to offer insights and stabilize the validity of key aspects of the evaluation protocols (\cite{protocol}). This is particularly true for classification tasks where the key features of a disease are not yet firmly established (matching the new-patterns discovery category; \cite{perspective}). 
\par The folding of the human cortical surface occurs during the perinatal period and remains constant for an individual for the rest of their life, much like a fingerprint, preserving early neurodevelopmental information (\cite{cachia}). Broadly, human brains share many features of cortical folding (sulci), however strong inter-individual variability creates inherent divergence in experts' opinions when it comes to labelling the more variable sulci, impeding the effort to automatically label sulci in the regions showing most variability.
\par While, to date, automated methods excel in the detection of most sulci, the variable shape and presence/absence of some sulci present a more complex computational hurdle (\cite{borne}). Successful automation through generalized, unbiased annotation would greatly aid studies focusing on brain folding variations, which are proxies for a critical developmental period with information that may relate to cognitive, behavioral, and developmental outcomes, along with psychiatric and neurological disorders. Brain folding is linked to brain function, and specific folding patterns correlate with susceptibility to neurological issues (\cite{jiang}). In particular, the morphology of the paracingulate sulcus (PCS), a highly variable sulcus, is associated with cognitive performance and hallucinations in schizophrenia (\cite{fornito}, \cite{garrison}, \cite{gay}, \cite{ath}). 
To this end, developing networks that can identify the presence or absence of PCS (or multiple PCS elements) in brain magnetic resonance imaging (MRI) and creating frameworks that can provide both local and global explanations would allow for a more systematic characterization and a more comprehensive understanding of whole-brain correlates related to its presence. This, in turn, would allow for both more precise and holistic assessments of its functional relevance and impact on brain functions (\cite{simons}).
\par However, a significant challenge arises in neuroimaging due to the high representational dimensionality of AI applications, the anatomical complexity of the brain, the use of unstable evaluation protocols, and the intricate yet uncertain nature of expert annotations. These factors make tasks in this domain particularly difficult to explain and evaluate. Conventional 2D local or global XAI methods often fall short and, in some cases, may even produce misleading explanations (\cite{perspective}). Additionally, as highlighted in \citep{mamalakis}, the issue of inter-method variability—where different XAI methods emphasize different features as important—can significantly erode trust in AI within the scientific community. These challenges underscore the urgent need for 3D explanation frameworks that can surpass the limitations of traditional XAI methods and provide more reliable global explanations.
\subsection{Aim and contribution of the study}
This study introduces, for the first time, a comprehensive framework to validate and explain deep learning models, providing transformative insights into pattern learning while establishing a new standard of credibility and reliability for such networks. The proposed XAI 3D-Framework tackles the intricate challenges of explainability in neuroimaging, enabling the detection and interpretation of complex patterns related to the presence or absence of the paracingulate sulcus (PCS), a highly variable cortical structure associated with reality monitoring and psychotic conditions. 
\par To achieve this, we developed an innovative methodology that employs two complementary local XAI methods, GradCam and SHAP, extended into 3D space to analyze the entire dataset used in the binary classification task. These methods are integrated with statistical patterns derived from the 3D brain inputs (Shape) via dimensionality reduction, producing global explanations that outperform traditional XAI methods in both interpretability and faithfulness. By combining these complementary approaches, our framework not only reveals the underlying learning patterns within the network but also significantly enhances the accuracy and clarity of the results. For the PCS classification task, we implemented two distinct 3D deep learning architectures: a simple 3D convolutional neural network (simple-3D-CNN) and a two-headed attention layer network with diverse backbone options (2CNN-3D-MHL and simple-3D-MHL). These networks utilized high-resolution 3D brain inputs, including grey-white surface boundaries and sulcal skeletons from both hemispheres. Leveraging a well-annotated cohort of 596 subjects from the TOP-OSLO study (\cite{morch-johnsen}), we trained, validated, and tested these networks, employing a 70\%-20\%-10\% data split.
\par Our framework introduces several pivotal innovations to the domain of explainable AI in neuroimaging: (i) By integrating statistical features (Shape) that are correlated with reduced dimensionality information, the framework ensures that the discovered patterns are not only grounded in the AI model's learning but also reflect cohort-specific variability. This dual-layered approach bridges statistical data and model-derived insights, enabling a deeper and more contextually relevant understanding of the results. (ii) The extension of established XAI methods, such as GradCam and SHAP, into the 3D domain addresses the critical need for 3D explanations in neuroimaging applications. Conventional 2D local or global XAI methods often fall short and, in some cases, may even produce misleading explanations. (iii) The use of GradCam and SHAP in tandem reduces inter-method variability and bolsters the reliability of the explanations, setting a new benchmark for trustworthy AI applications. The proposed multi-method framework delivers robust and actionable insights, particularly in complex neuroimaging tasks such as cortical morphology studies.
\par Notably, the XAI 3D-Framework demonstrated superior performance compared to traditional XAI methods in terms of faithfulness for global explanations, successfully identifying significant sub-regions of an atlas brain (the ICBM 2009a Nonlinear Asymmetric atlas, \cite{atlas1,atlas2}). This capability provides a transparent and reliable mechanism to trace the patterns driving network decisions, enhancing trust and enabling deeper exploration of deep learning model outputs in neuroscience. By combining methodological rigor with practical innovation, this framework opens new avenues for understanding and interpreting brain structure-function relationships, making it a foundational tool for advancing both research and clinical applications in neuroimaging.
\subsection{Related work}
\subsubsection{The application: Sulcal pattern studies}
\par Cortical folding, which develops during the perinatal period (i.e., in the last few months of gestation and the first few months after birth), results in significant inter-individual variability often overlooked in population studies. Understanding the variability of sulcal patterns is critical for multiple reasons: strict descriptive anatomy, refinement of inter-subject registration, investigation of neurodevelopmental mechanisms, and the search for anatomo-functional correlates. These patterns hold potential for investigating healthy functional variability (e.g., the relationship between cingulate folding patterns and functional connectivity \cite{fedeli}) and pathological outcomes (e.g., paracingulate folding linked to hallucinations in schizophrenia \cite{rollins}).
\par While the study of cortical folding variability has been approached through global methods—considering whole-brain or regional sulcal parameters such as gyrification index or sulcal pits—finer investigations often require a focus on specific sulci. This underscores the need for automated sulcal recognition methods. Although several techniques have been developed for general sulcal labeling (reviewed in \cite{mangin}), to the best of our knowledge, no current method can automatically label the PCS in a 3D approach; most rely on 2D analyses of specific MRI slices (\cite{yang}). The omission of PCS in whole-brain labeling stems from its complex anatomical specification, defined not only by its location but also by its orientation (parallel to the cingulate sulcus). Consequently, even newer labeling frameworks fail to identify the PCS within the medial frontal cortex (\cite{borne}).
\par This gap is particularly critical given the potential importance of investigating whole-brain anatomical correlates of PCS variability in understanding severe symptoms of psychosis. Automatic detection of the PCS is valuable for large dataset exploration, especially given its links to functional variability in reality monitoring (\cite{simons}). Moreover, incorporating an explainability component through advanced XAI frameworks is unprecedented and transformative. It allows not only for the identification of the PCS but also for uncovering new patterns in its anatomical covariates, offering insights into the functional mechanisms underlying its role. The integration of XAI with AI classification thus provides a robust platform for both enhancing interpretability and advancing our understanding of the broader neurodevelopmental and pathological contexts associated with the PCS.
\subsubsection{Explainable methods}
Recent studies have seen a significant surge in the exploration of XAI within medical image analysis and neuroimaging domains (\cite{xai,mx1,mx2,mine}). XAI methodologies are broadly categorized into interpretable and post-hoc approaches. Interpretable methods focus on models that possess inherent properties such as simulatability, decomposability, and transparency, often linked to linear techniques like Bayesian classifiers, support vector machines, decision trees, and K-nearest neighbor algorithms (\cite{xai22}). On the other hand, post-hoc methods are typically used with AI techniques to reveal nonlinear mappings within complex datasets (\cite{xai,xai22}).
\par A widely used post-hoc technique is Local Interpretable Model-Agnostic Explanations (LIME), which explains the network's predictions by building simple interpretable models that approximate the deep network locally, i.e. in the close neighborhood of the detected structure (\cite{xaimi}). Post-hoc techniques include both model-specific approaches that address specific nonlinear behaviors and model-agnostic approaches that explore data complexity (\cite{xai,xai22}). In computer vision, model-agnostic methods such as LIME and perturbation techniques are widely used, while model-specific methods encompass feature relevance, condition-based explanations, and rule-based learning (\cite{lime,xaims,xai22}). 
\par In medical imaging, explainable methods often focus on attribution and perturbation techniques (\cite{xai_surv}).
Attribution techniques like LIME as well as Layer-wise Relevance Propagation (LRP), Gradient-weighted Class Activation Mapping (GradCAM), and Shapley Additive Explanations (SHAP) identify important features for a given prediction by assigning relevance scores to the input features. Perturbation techniques assess the sensitivity of an AI prediction to specific input features by systematically altering sub-groups of the input data (\cite{xai_surv,xaimi}). GradCAM is notably prevalent among explainable methods in medical imaging due to its ease of application and understanding, as well as its ability to map significant features in the imaging space using the activations of the last convolutional layers (\cite{xaimi}).
\par By advancing these explainable methodologies we can better interpret complex models, enhancing their transparency and trustworthiness, particularly in applications like medical imaging, and neuroimaging. An important gap in the literature is the absence of three-dimensional representation frameworks that can explain complex models, such as those used in neuroscience, by providing faithful global explanations. Such frameworks could offer more accurate interpretations than established approaches, potentially improving AI transparency and uncovering new patterns in significant sub-regions involved in classification and prediction.
\section{Methods and Background}
In this study, we utilized different 3D classifiers to address the binary classification problem of determining the presence or absence of PCS. Furthermore, we developed a novel XAI 3D framework for non-explainable and new-pattern discovery tasks (\cite{perspective}). We employed two distinct explanation methods from the post-hoc family, SHAP and GradCam, which were expanded into three-dimensional space. The outcomes from these two explainable methods (SHAP and GradCam) were concatenated with results derived from a statistical feature extraction model (Shape). The statistical model is a transparent dimensionality reduction algorithm applied to the input data of the validation cohort. This comprehensive approach aims to mitigate potential biases and enhance the robustness of pattern learning discovery. 
\subsection{Architectural design of deep learning networks for the classification task}
\begin{figure*}
 \medskip
\relax \textbf{a}
\centerline{
    \includegraphics[trim={0.25cm 0.05cm 0.05cm 0.05cm},clip,scale=.35]{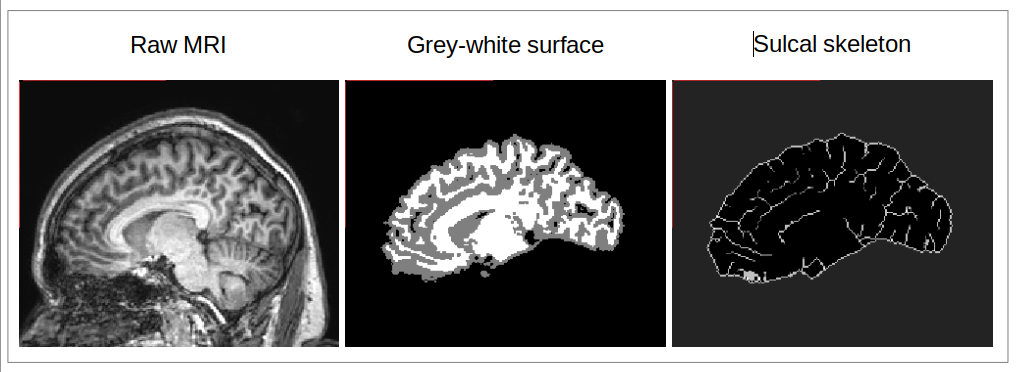}}
    \relax \textbf{b}
\centerline{
    \includegraphics[scale=.425]{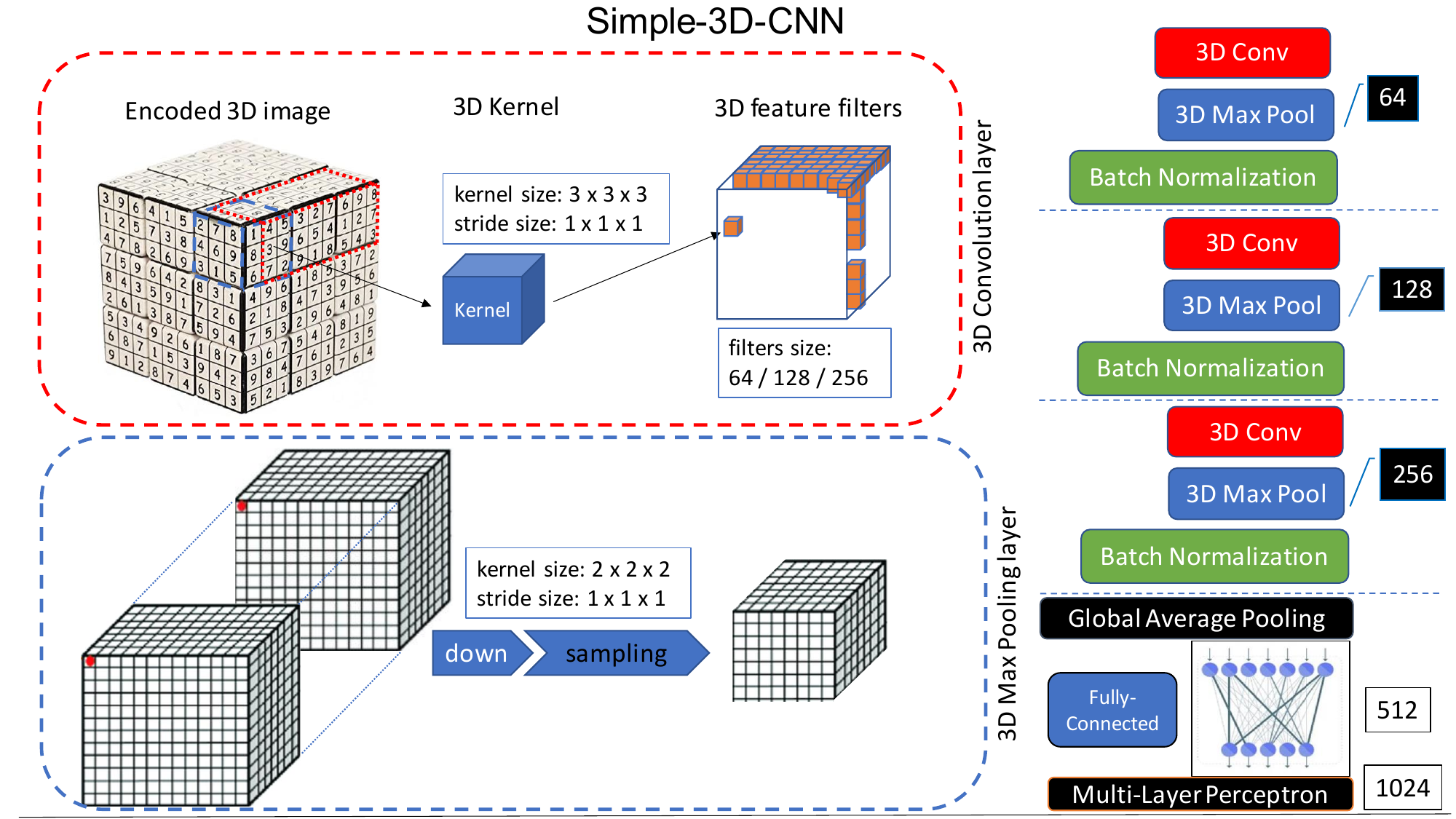}
}
    \relax \textbf{c}
    \centerline{

    \includegraphics[scale=.425]{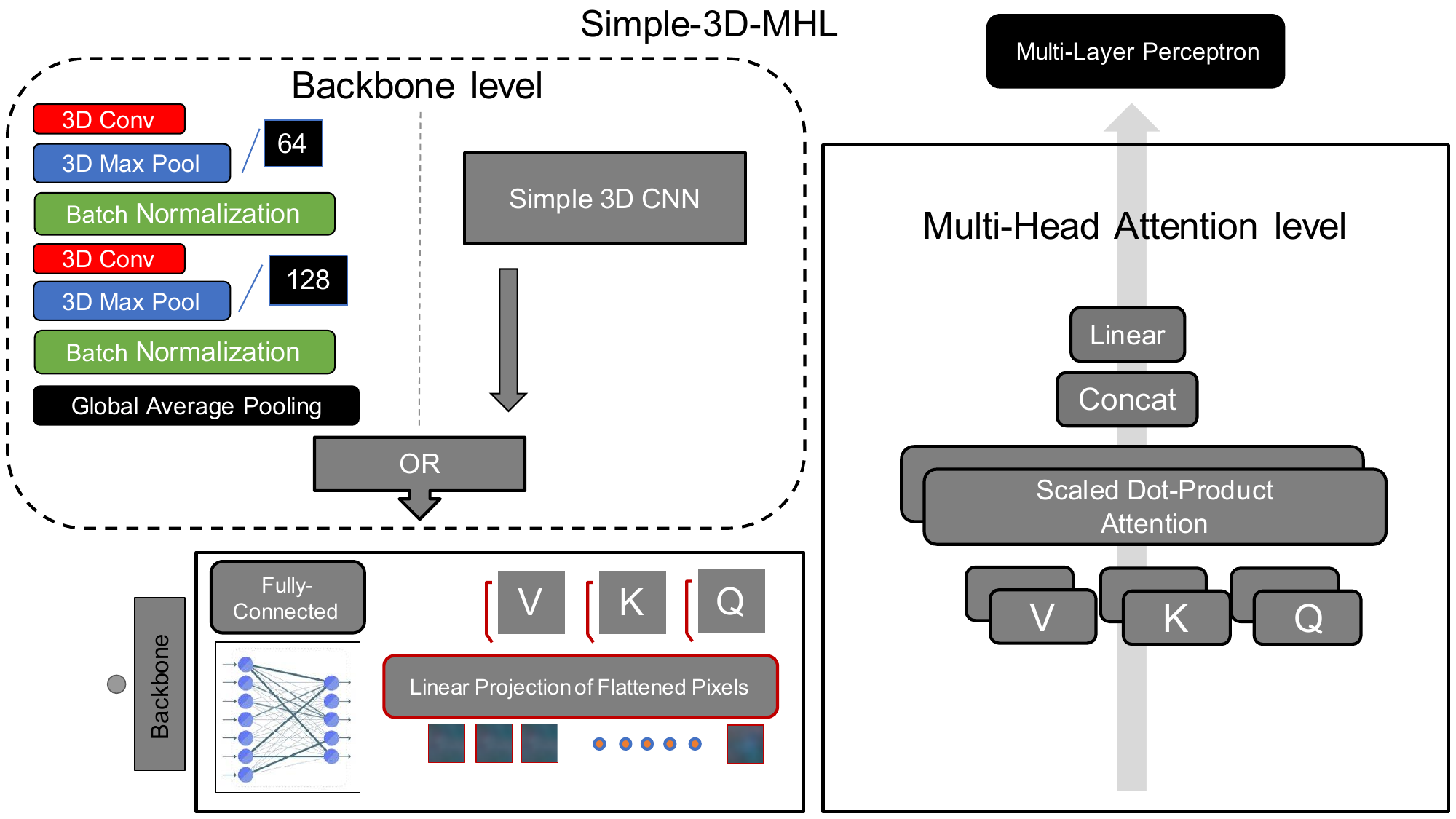}
}
\caption{The input modalities and the architecture of simple-3D-CNN and the simple-3D-MHL networks.
 \textbf{a,} Input modalities illustrated on a right hemisphere coronal slice. In more detail, the raw MRI of a given subject, the corresponding grey-white surface, and the corresponding sulcal skeleton. \textbf{b,} The architecture of simple-3D-CNN network with explanation of 3D Convolution layer (3D Conv) and 3D Max Pooling layer on the left. \textbf{c,} The three dimension MHL model with two different backbone choices, the full simple-3D-CNN (simple-3D-MHL) and the two level simple-3D-CNN layer (2CNN-3D-MHL).}
    \label{ml1}
\end{figure*}

We used two different deep learning models for the binary classification task (Fig. \ref{ml1}b.,c.). The first network was a 3D Convolutional Neural Network (CNN) with five levels (see Fig. \ref{ml1}b.). Each level incorporated a CNN block with a 3D convolution layer, a 3D max-pooling layer, and a batch normalization layer. In the first three levels, the 3D convolution layer employed 64, 128, and 256 filters, respectively (as shown in Fig. \ref{ml1}b.). The last level connected to a multi-layer perceptron (MLP) for the final prediction (as illustrated in Fig. \ref{ml1}a.). The MLP comprised three distinct perceptrons and two dropout layers to estimate epistemic uncertainty.
For the second network (Fig. \ref{ml1}c.), we used a combination of multi-head attention layers (MHL) to focus on the global diversity and variation of a backbone output. To reduce the biased choice of only one backbone selection, we opted for two distinct backbone networks: (i) a 3D Convolution layer block with two levels of 64 and 128 filters a Global Average Pooling, and a perceptron of 32400 hidden layers (2CNN-3D-MHL), and (ii) the straightforward 3D CNN network outlined previously (Fig. \ref{ml1}b., simple-3D-MHL). We used the multi-head attention mechanism described in (\cite{mha}) and as we used a two-head attention of the same input ("self-attention").
The two heads ($head_c$) are given by:
\begin{equation}
head_c=AT(QW^Q_c,KW^K_c, VW^V_c)
\label{14}
\end{equation}
where $c$ is the backbone output. There are two outputs corresponding to the two uses of the backbone, and each output is connected to a perceptron with $N$ hidden layers (where $N$ is equal to the product of the weight and height of the input image). The $AT$ is the attention layer and it computed by:
\begin{equation}
AT(Q,K,V)= softmax(\frac{QK^T}{\sqrt{d_k}})V
\label{15}
\end{equation}
where the input matrix are combinations of queries and keys of dimension $d_k$, and values of dimension $d_v$. Queries are packed together into a matrix $Q$. The keys and values are also packed together into matrices $K$ and $V$. The output of the MHL network is given by:
\begin{equation}
MHL=Concat(head_c,head_c)
\label{121}
\end{equation}
where $c$ defines by the $backbone_{output}$.
The output of the MHL was passed again from the MLP presented above to make the final prediction.
\subsection{Extending explainability methods in 3D space}
We used two distinct explainable techniques: the widely-utilized sensitivity local explainability technique in medical imaging applications known as the GradCam method ((\cite{gcam}), and a robust attribution explainability technique called SHAP ((\cite{shap}).
\par The significance of the 3D space in neuroscience applications emphasises a necessity to extend 2D XAI methods into three dimensions. In the pursuit of computing the class-discriminative localization map encompassing width $w$, height $h$, and depth $d$ within a specific 3D brain MRI corresponding to a class $c$ (PCS or noPCS), the computation involves determining the gradient of the score for class $c$, denoted as $y^c$, in relation to the $n^{th}$ feature activation map ($A^n$) of the final convolution layer in each deep network. To determine the importance weights ($\alpha^c_n$) for each $n$ feature activation map, global average pooling is employed over the width ($i$), height ($j$), and depth ($k$) of each feature.
\begin{equation}
\alpha^c_n=\frac{1}{Z}\sum_i{{\sum_j\sum_k\frac{dy^c}{dA_{ijk}^n}}}
\label{20}
\end{equation}
where $Z$ the summation of $i$, $j$ and $k$.
Moreover, we used a weighted combination of forward activation maps and a $ReLU$ to deliver the final GradCam activation map.
\begin{equation}
GradCam=ReLU(\sum_n{\alpha^c_n A_{ijk}^n})
\label{21}
\end{equation}
SHAP computes the attribution of each pixel of an input image for a specific prediction of a computer vision task. Attribution explainability methods follow the definition of additive feature attribution mainly as a linear function of:
\begin{equation}
g(f,x)=\phi_0 + \sum_{i=1}^M\phi_i x_i
\label{22}
\end{equation}
where $f$ is the prediction network, $g(f,x)$ is the explanation model, $\phi_i$ is the importance of each feature attribution ($\phi_i \in \mathbb{R}$), and $M$ is the number of simplified input features (pixels). Shapley value estimation is one of the main mathematical formulations that the SHAP algorithm uses to assign an importance value to each feature, representing the effect on the model prediction of including that feature (attribution). If we define a subset $S$ of the total feature space ($F$) of an input 3D image ($i=1...N$, where $N$ is the number of samples in the dataset), and $x_i$ is a 3D matrix of width $w$ and height $h$ and depth $d$ for the $i^{th}$ sample, and $x_S$ is the subset of chosen features in the 3D space, then:
\begin{equation}
\phi_i=\sum_{S/(i)} \frac{|S|!(|F|-|S|-1)!}{|F|!} [f_{S(i)}(x_{S(i)})-f_S(x_S)]
\label{23}
\end{equation}
Here, $f_{S(i)}$ is a model trained with the presented $x_s$ features, and $f_S$ is another model trained with the features withheld. For our study, we used Deep SHAP (\cite{shap} to describe our deep learning network models. This approach uses a chain rule and linear approximation as described in (\cite{shap}).
\subsection{The XAI 3D-Framework}
The XAI 3D-Framework introduces a groundbreaking approach to generating global explanations that facilitate the discovery of new patterns in neuroimaging studies. Our proposed framework uniquely integrates statistical features derived from cohort data (Shape) with insights from two complementary explainability methods, GradCam and SHAP, in a three-dimensional space. By combining statistical information with model-driven learning, the framework provides a dual-layer understanding: one rooted in the cohort's inherent variability and the other reflecting the AI model's decision-making processes based on the classification task. Such integration not only enhances robustness but also minimizes inter-method variability and potential biases (\cite{mamalakis}), ensuring reliable and interpretable outcomes. The use of multiple methods, as opposed to relying on a single explainability approach, represents a significant step forward in delivering more comprehensive and trustworthy explanations, especially in challenging contexts like cortical morphology.
\par
\begin{figure*}
      \relax \textbf{a}
\centerline{
    \includegraphics[trim={1.85cm 0.80cm 1.cm 0.8cm},clip,scale=.55]{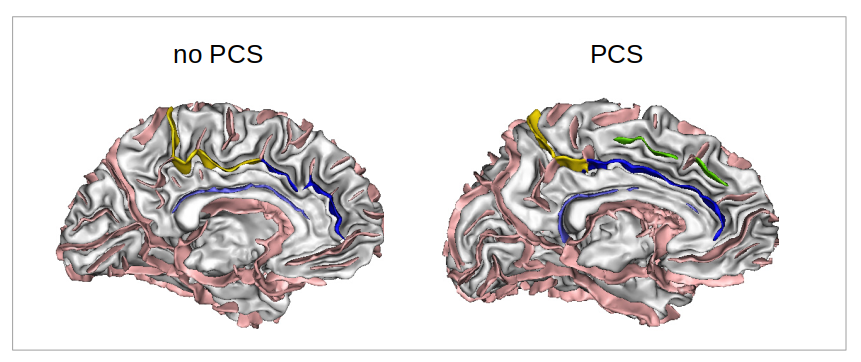}}
    \relax \textbf{b}
    \centerline{
   \includegraphics[trim={0.05cm 0.0cm 0.05cm 0.0cm},clip,scale=.535]{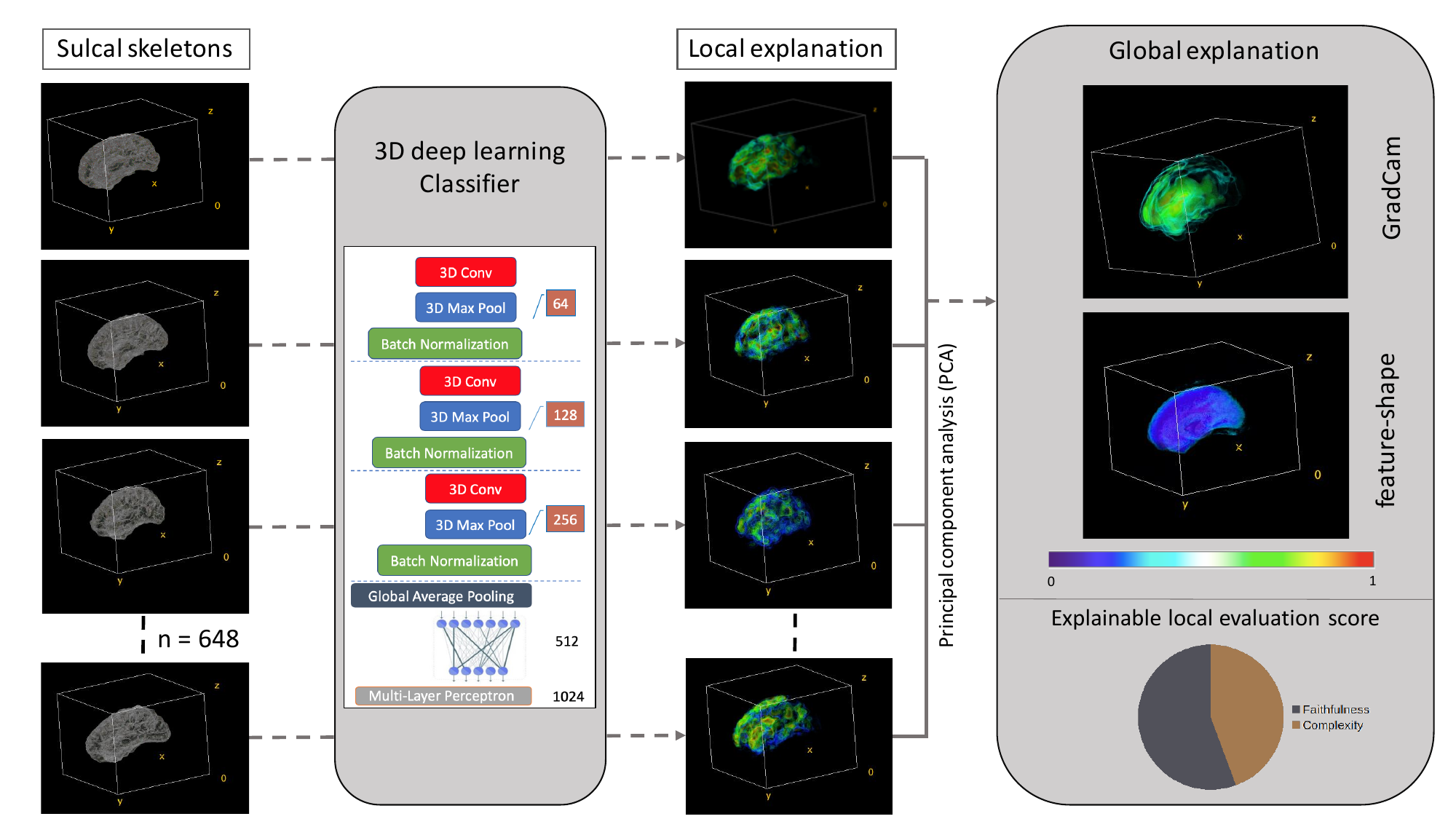}}
   \caption{Classification task determination and the transition from local to global 3D explaination.
\textbf{a,} Illustration of the no PCS condition (no PCS), and the PCS condition (PCS, in green line) on two left hemisphere 3D white matter reconstructions obtained with BrainVISA. The cingulate sulcus is coloured in yellow and blue and the callosal sulcus is coloured in purple. \textbf{b,} The 3D explainable framework that provides both local and global interpretations and explanations of our deep learning 3D classification network's results. The ratio of the faithfulness and complexity metrics were computed at that stage. In this example we include only the GradCam explainability method  for simplicity.}
  \label{pcs}
\end{figure*}

\begin{figure}[ht!]
\centerline{
   \includegraphics[trim={0.05cm 0.0cm 0.05cm 0.0cm},clip,scale=.55]{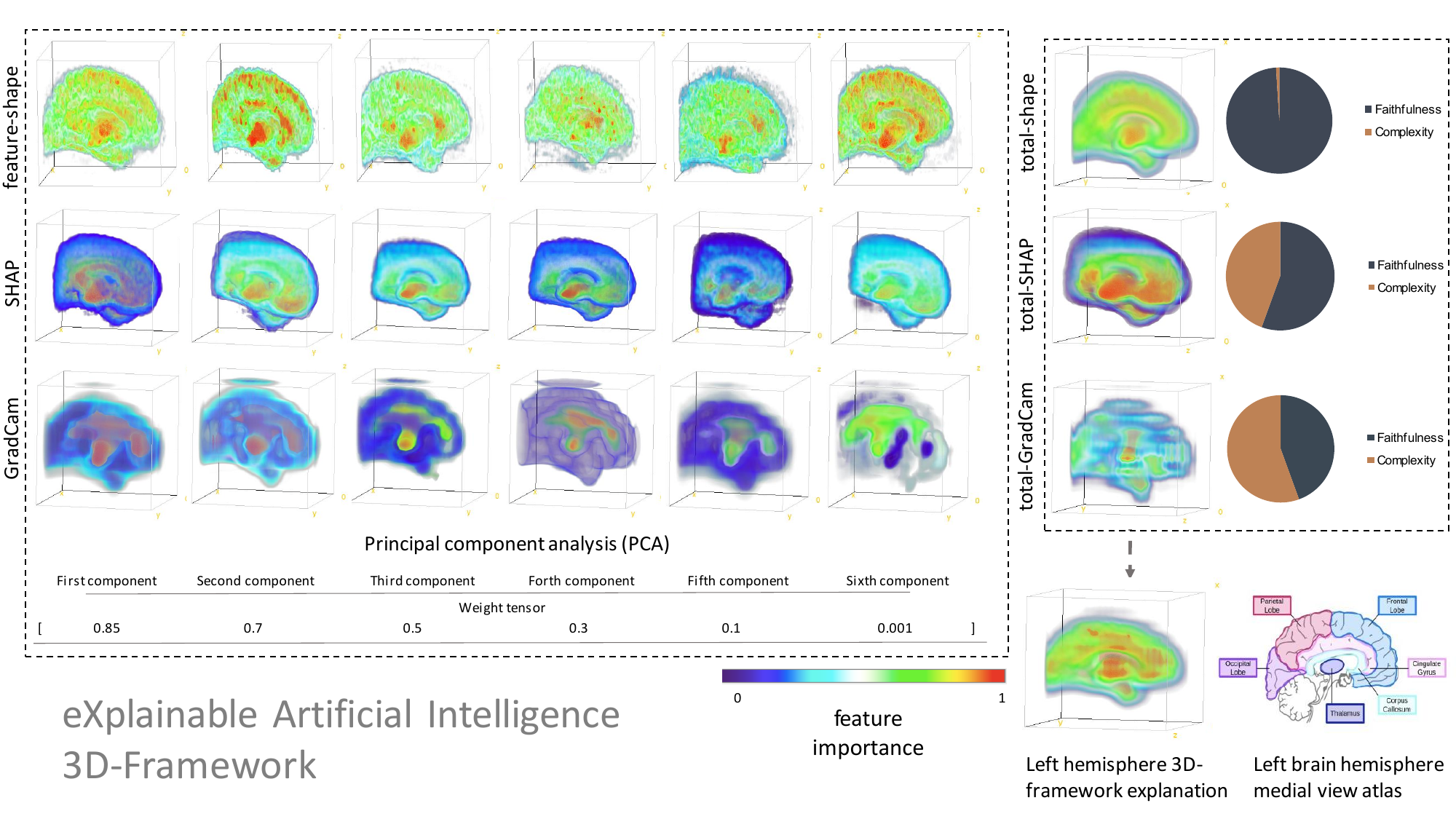}} 
   \caption{The proposed global 3D-Framework explanation.
 A weighted averaging (Weight tensor: [0.85, 0.7, 0.5, 0.3, 0.1, 0.001]) of six PCA components produces the average PCA image for PCA-Shape, PCA-GradCam, and PCA-SHAP. Following this, a weighted averaging (Weight tensor: [0.85, 0.5, 0.1]) of the three Global PCA overlapping images extracted the total overlapping image. This total overlapping image was then registered on a sulcal probabilistic atlas (the ICBM 2009a Nonlinear Asymmetric atlas, \cite{atlas1,atlas2}) to unveil the model's pattern for determining the presence or absence of the PCS.}
  \label{pcs22}
\end{figure}

 To achieve its goals, the framework employs GradCam and SHAP, two widely used explainability methods, expanded into three-dimensional space to extract local explanations from deep learning classifiers (simple-3D-CNN and simple-3D-MHL; Fig. \ref{pcs} b.). Faithfulness and complexity scores are assigned to evaluate the quality of these local explanations, ensuring they meet predefined thresholds (see Supplementary Material Table 2). To generate global explanations, we apply Principal Component Analysis (PCA; (\cite{pca})) to both the sulcal skeleton and grey-white surface 3D brain inputs, capturing a variety of global feature importance patterns (PCA-Shape; see Fig. \ref{pcs22}). After testing different configurations, the optimal solution—capturing over 80\% of the cohort variance—was achieved with six PCA components.
\par The global explanations derived from GradCam and SHAP yielded superior faithfulness and complexity results when averaged across the six PCA components using a weighted tensor: [0.85, 0.7, 0.5, 0.3, 0.1, 0.001] (see Supplementary Material Table 2; PCA0-SHAP vs. total-SHAP, and PCA0-GradCam vs. total-GradCam). These weighted averages formed the final global explanations (total-GradCam and total-SHAP). In parallel, the same approach was applied in the Shape domain to determine the global statistical feature importance of sulcal skeleton and grey-white surface inputs (total-Shape; see Fig. \ref{pcs22}).
\par By integrating statistical and model-driven insights, the framework provides a robust platform for uncovering meaningful patterns while maintaining the reliability and interpretability essential for applications in neuroimaging.
The weighted average formulation is defined as follows: 
\begin{equation}
G(X,W)= \frac{\sum{w_i x_i}}{\sum{w_i}}
\label{1212}
\end{equation}
where W is the weight tensor and the X is the pixel images tensor.
\subsection{The global explanation of the 3D-Framework}
To compute the global explanation of the 3D framework, we first manually aligned and rescaled the two global XAI explanations (total-SHAP, total-GradCam) to the average of the six components of the PCA-Shape (total-Shape). Then, we used equation \ref{1212}, incorporating a three-component weight tensor of $[0.85, 0.5, 0.1]$. The weight combination [0.85, 0.5, 0.1] was chosen after empirical testing of multiple combinations, to optimize faithfulness while maintaining reasonable complexity scores. Preliminary results showed that higher weights for total-Shape (e.g., 0.85) yielded superior faithfulness scores, reflecting the importance of feature importance in global explanations. This selection balances the trade-offs between faithfulness, complexity, and redundancy across the metrics. After we define the best combination, we conducted an ablation study to identify the optimal combination of the weight tensor. The different cases we examined were denoted as 851, 815, 185, 158, 518, and 581, which are fixed-order representations of the methods in the following sequence: total-Shape, total-SHAP, and total-GradCam explanations. The weights correspond to 0.85 as '8', 0.5 as '5', and 0.1 as '1'. For example, 3D-Framework-851 refers to the proposed 3D-Framework with weight values of 0.85 for total-Shape, 0.5 for total-SHAP, and 0.1 for total-GradCam. These components were derived from both the sulcal skeleton and the grey-white surface inputs of the total dataset, considering both the right and left hemispheres. In our specific study, the best combinations identified were 851 and 815 (see Supplementary Material Table 2). 
\par To minimize potential bias that might occur by relying on only one deep learning network, we applied this approach to both the simple-3D-CNN and simple-3D-MHL networks.
Identifying the significant features of the networks explanations provides insights into the mechanisms driving the network's decision-making process. For enhanced clarity, the key brain sub-regions of interest, corresponding to the sulcal skeleton and grey-white surface, are visually depicted in Fig. \ref{ml1}a. Finally, the global explanation from the 3D framework was registered on a sulcal probabilistic atlas to illustrate the model's pattern associated with determining the presence or absence of the PCS. The registration process involved affine transformations, including translation, rotation, scaling, and geometry adjustments, to align with the probabilistic atlas (the ICBM 2009a Nonlinear Asymmetric atlas, \cite{atlas1,atlas2}).
\subsection{Cohort's description and pre-processing image analysis}
We used the structural MRI of 596 participants from the TOP-OSLO study (\cite{morch-johnsen}) for a binary classification task. The participants encompassed individuals with a diagnosis on the schizophrenia spectrum (183), on the bipolar disorder spectrum (151), and unaffected control participants (262). T1-weighted images were acquired using a 1.5 T Siemens Magnetom Sonata scanner (Siemens Medical Solutions, Erlangen, Germany). 
\par Two experts, A.A. and H.V., performed image annotations, categorizing them into two classes: 'no paracingulate sulcus' (noPCS) and 'paracingulate sulcus' (PCS). This annotation was based on the protocol described in (\cite{protocol}) with further details available in the supplementary materials under '1.1 PCS classification of TOP-OSLO' and is illustrated in Fig. \ref{pcs}a. A more analytical description of the TOP-OSLO cohort details is available in the study \cite{morch-johnsen}.
\subsection{Processing of two distinct images inputs}
We initially processed the brain structural MRIs using the BrainVISA software (\cite{brainvisa}) extracting two images as inputs for the classifier: the grey-white surface and the sulcal skeleton. These were extracted from the raw MRI with an established protocol consisting of bias correction, histogram analysis, brain segmentation, hemisphere separation, dichotomization of the white matter from the union of grey matter and cerebrospinal fluid, and skeletonization of the result, as detailed in (\cite{riv}). Specifically, the grey-white surface was obtained by minimizing a Markov field and the segmentation used homotopic deformations of the hemisphere bounding box, resulting in the grey-white surface, where voxels are dichotomised into either grey or white.  The skeleton was then derived from this object by applying a homotopic erosion embedding a watershed alogrithm that preserves the initial topology resulting in the sulcal skeleton. These two modalities were then used to train and evaluate our networks as well as for our explainability methods. Fig. \ref{ml1}a,b,c. show the structural MRI and the corresponding grey-white surface and sulcal skeleton outputs from BrainVISA.
\subsection{Hyper-parameter initialization}
After randomly shuffling the data, each dataset was split into training, validation, and testing sets containing $70\%$, $20\%$, and $10\%$ of the total number of images, respectively. Sparse categorical cross-entropy was used as the cost function and the loss function were optimized using the Adam algorithm (\cite{adam}). After manual hyper-parameter searching the best learning rate was a value of $0.0001$. We utilized a strategy of exponentially decreased during the first 50 epochs and then fixed at $0.0001$ for the last 50 epochs. To train the networks, an early stopping criterion of 10 consecutive epochs was employed and a maximum of 100 epochs was used for both input modalities (sulcal skeleton and grey-white surface) for both the left (L) and right (R) hemispheres. Finally, we use data augmentation techniques including rotation (around the center of the image by a random angle from the range $[-15^{\circ}, 15^{\circ}]$), width shift (up to $20$~pixels), height shift (up to $20$~pixels), and Zero phase Component Analysis (ZCA; \cite{noise}) whitening (add noise in each image) to avoid overfitting.
\subsection{Evaluation metrics for the explanation}
A crucial aspect of this study lies in evaluating how accurate and comprehensive were the local and global explanations. To derive a useful explanation, two primary scores play a pivotal role: faithfulness and complexity. An intuitive way to assess the quality of an explanation is by measuring its ability to accurately capture how the predictive model reacts to random perturbations (\cite{fid}). For a deep neural network $f$, and input features $\boldsymbol{x}$, the feature importance scores (also known as "attribution scores") were derived in a way such that when we set particular input features $\boldsymbol{x_s}$ to a baseline value $\boldsymbol{x_s^f}$, the change in the network's output was proportional to the sum of attribution scores of the perturbed features $\boldsymbol{x_s}$. We quantified this by using the Pearson correlation between the sum of the attributions of $\boldsymbol{x_s}$ and the difference in the output when setting those features to a reference baseline (\cite{f}). Thus, we define the faithfulness of an explanation method $g$ as:
\begin{equation}
M_{faith}(f,g;\boldsymbol{x}) = corr_S(\sum_{i\in S} g(f,\boldsymbol{x})_i,f(\boldsymbol{x})-f(\boldsymbol{x}[\boldsymbol{x_s}=\boldsymbol{x_s^f}]))
\label{1}
\end{equation}
where $S$ is a subset of indices ($S\subseteq [1,2,3 ... d]$), $\boldsymbol{x_s}$ is a sub-vector of an input $\boldsymbol{x}$ ($\boldsymbol{x}=\boldsymbol{x_s}\cup \boldsymbol{x_f}$ and $\boldsymbol{x_f}$ the unchanged features of $\boldsymbol{x}$ image). The total number of the $\boldsymbol{x_s}$ sub-vectors, which partition an image is $d$. We denote as $\boldsymbol{x_s}$ the changed features, and $\boldsymbol{x_f}$ the unchanged features of $\boldsymbol{x}$ image. 
\par If for an image $\boldsymbol{x}$, explanation $g$ highlights all $d$ features, then it may be less comprehensible and more complex than needed (especially if $d$ is large). It is important to compute the level of complexity, as an efficient explanation has to be maximally comprehensible (\cite{f}). 
If $P_g$ is a valid probability distribution and the $P_g(i)$ is the fractional contribution of feature $x_i$ to the total magnitude of the attribution, then we define the complexity of the explanation $g$ for the network $f$ as:
\begin{equation}
M_{compx}(f,g;\boldsymbol{x}) = \sum_{i=1}^{d}P_g(i)(log(\frac{1}{P_g(i)}))
\label{4}
\end{equation}
where:
\begin{equation}
P_g(i) = \frac {\vert g(f,\boldsymbol{x})_{i}\vert}{\sum_{j=1}^{d}\vert g(f,\boldsymbol{x})_{j}\vert}
\label{5}
\end{equation}
In order to evaluate these two explainability metrics, we used the software developed by (\cite{quan}. This software package is a comprehensive toolkit that collects, organizes, and evaluates a wide range of performance metrics, proposed for explanation methods. Note that we used a zero baseline ('black'; $\boldsymbol{x_s^f}$= $\boldsymbol{0}$) and 70 random perturbations to calculate the faithfulness score.
\par Finally, to extract the faithfulness and complexity scores of the global explanations for the total-SHAP, total-GradCam, and the proposed 3D-Framework, we utilized again the software developed by (\cite{quan}). As a first step we manually aligned and rescaled the two global XAI explanations (total-SHAP and total-GradCam) to the total-Shape. The input image consisted of the total-Shape results, while the total-GradCam, total-SHAP, and 3D-Framework global explanation served as reference explanations, respectively. In this context, the score of a global explanation makes sense, as the individual input brains of the cohort were aligned in the same template and the most significant variability of each class was assigned in the total-Shape. Consequently, we anticipated the classifier to classify correctly whether a MRI has a PCS or not.  
\section{Results}
\subsection{Classifiers performance for presence or absence of paracingulate sulcus}
For brevity, we discuss in the main manuscript only the simple-3D-MHL results which outperformed the two-level CNN backbone network (2CNN-3D-MHL). The analytical tables and results for 2CNN-3D-MHL can be found in the supplementary material subsection 2.1 ('Classification results of the 2CNN-3D-MHL' ; Table 1).
\par The performance of the simple-3D-MHL network in the left hemisphere was higher (around 73.00\% in all testing metrics and 74.10\% in all validation metrics) than that in the right hemisphere (around 58.00\% in all testing metrics and 63.10\% in the validation metrics). For the simple-3D-CNN, the performance of the network in the left hemisphere was higher (around 72.90\% in all testing metrics and 74.00\% in all validation metrics) than that in the right hemisphere (around 56.00\% in all testing metrics and 63.00\% in the validation metrics). 
The analytical Figures and outcomes for the simple-3D-MHL and simple-3D-CNN networks are presented in supplementary material subsection 2.2 ('Additional global explainability methods and different components PCA results'; Supplementary  Fig. 1 a.,b.).
\par The discrepancy in performance between the left and right hemispheres was to be expected for two reasons. First, the PCS is more prominent in the left than in the right hemisphere (\cite{paus_human_1996, yucel_hemispheric_2001}), including in psychopathological conditions such as schizophrenia (\cite{garrison}). Furthermore, the left PCS has a greater number of associations with regional cortical thickness and sulcal depth than the right PCS (\cite{fornito}), implying more covariability of anatomical features contained in either of our input modalities with the presence of the PCS in the left hemisphere than the right.
\subsection{Global explanations and their PCA component results}
We extracted and evaluated the explainability results from both networks to avoid biased observations and to investigate whether there was a clear cause-and-effect relationship between the quality of explanation and prediction performance. We present the results of the simple-MHL network in the main manuscript as it had slightly better performance compared to the simple-3D-CNN. The results for the simple-3D-CNN are thoroughly detailed in the supplementary material subsection 2.2 (refer to 'Additional global explainability methods and different components PCA results'; Supplementary  Fig. 2 and 3). 
\par
\begin{figure}
      \medskip     
\centerline{
\relax \textbf{a}
    \includegraphics[trim={0.0cm 0.0cm 0.0cm 0.0cm},clip,scale=.5]{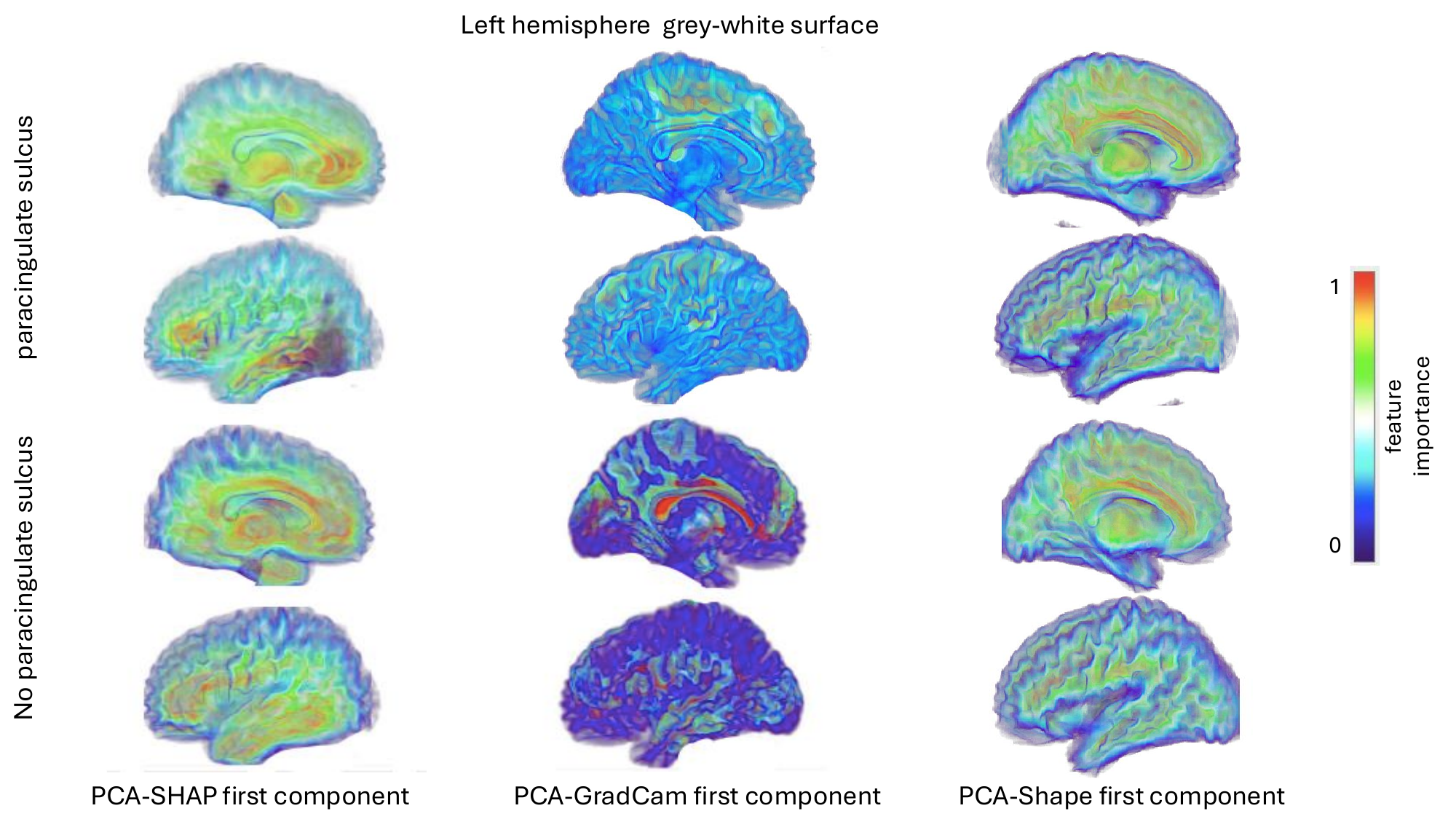}}
\centerline{
\relax \textbf{b}
    \includegraphics[trim={0.0cm 0.0cm 0.0cm 0.0cm},clip,scale=.5]{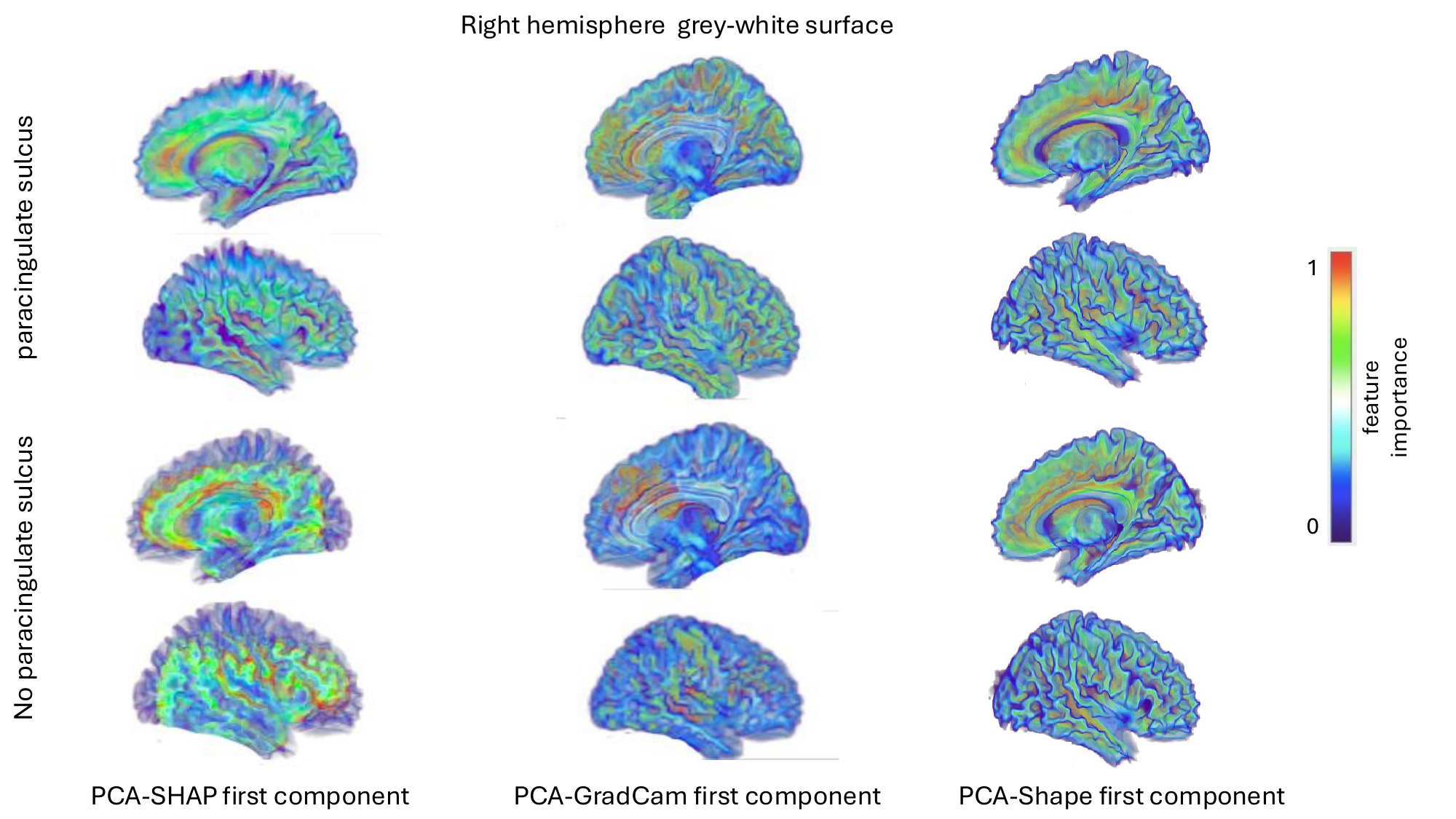}}
\caption{Simple-3D-MHL results on the left and right hemisphere of grey-white surface brain inputs.
\textbf{a,b,} show the explainability results for the PCS class images of the first component among the six components of PCA for the total input modality (PCA-Shape), the total corresponding GradCam results (PCA-GradCam), and the total corresponding SHAP results (PCA-SHAP).  The feature's importance (pixel attribution) varies from 0 (blue color) to 1 (red color), with high importance being 1 for the PCA-GradCam and PCA-Shape results. The orientation of the results are based on the medial anatomical views. All the presented results are align and mapping in the ICBM 2009a Nonlinear Asymmetric atlas (\cite{atlas1,atlas2}).}
  \label{x2}
\end{figure}

 The first component analysis of PCA explainability results of simple-3D-MHL networks on the left and right hemisphere of the grey-white surface inputs mainly focus on the frontal lobe (mostly inferior lateral and inferior medial), the cingulate gyrus, the temporal lobe, and occasionally the thalamus for detecting the presence or absence of the PCS. More  specifically, for the detection of the presence of PCS (paracingulate sulcus; Fig. \ref{x2}) in the left and right hemisphere the simple-MHL network focuses more in the frontal lobe (medial and inferior lateral), cingulate gyrus (mostly anterior), temporal lobe and sometimes thalamus. Conversley, for the absence of PCS (No paracingulate sulcus; Fig. \ref{x2}) in the left hemisphere the network focuses in the frontal lobe (mostly inferior medial and inferior frontal), the temporal lobe, the cingulate gyrus, occasionally the thalamus, and specifically for the PCA-GradCam, the corpus callosum. On the right hemisphere the simple-MHL network focuses in frontal lobe (medial and lateral), the temporal lobe, and the cingulate gyrus.
\par
\begin{figure}
      \medskip     
\centerline{
\relax \textbf{a}
    \includegraphics[trim={0.0cm 0.0cm 0.0cm 0.0cm},clip,scale=.475]{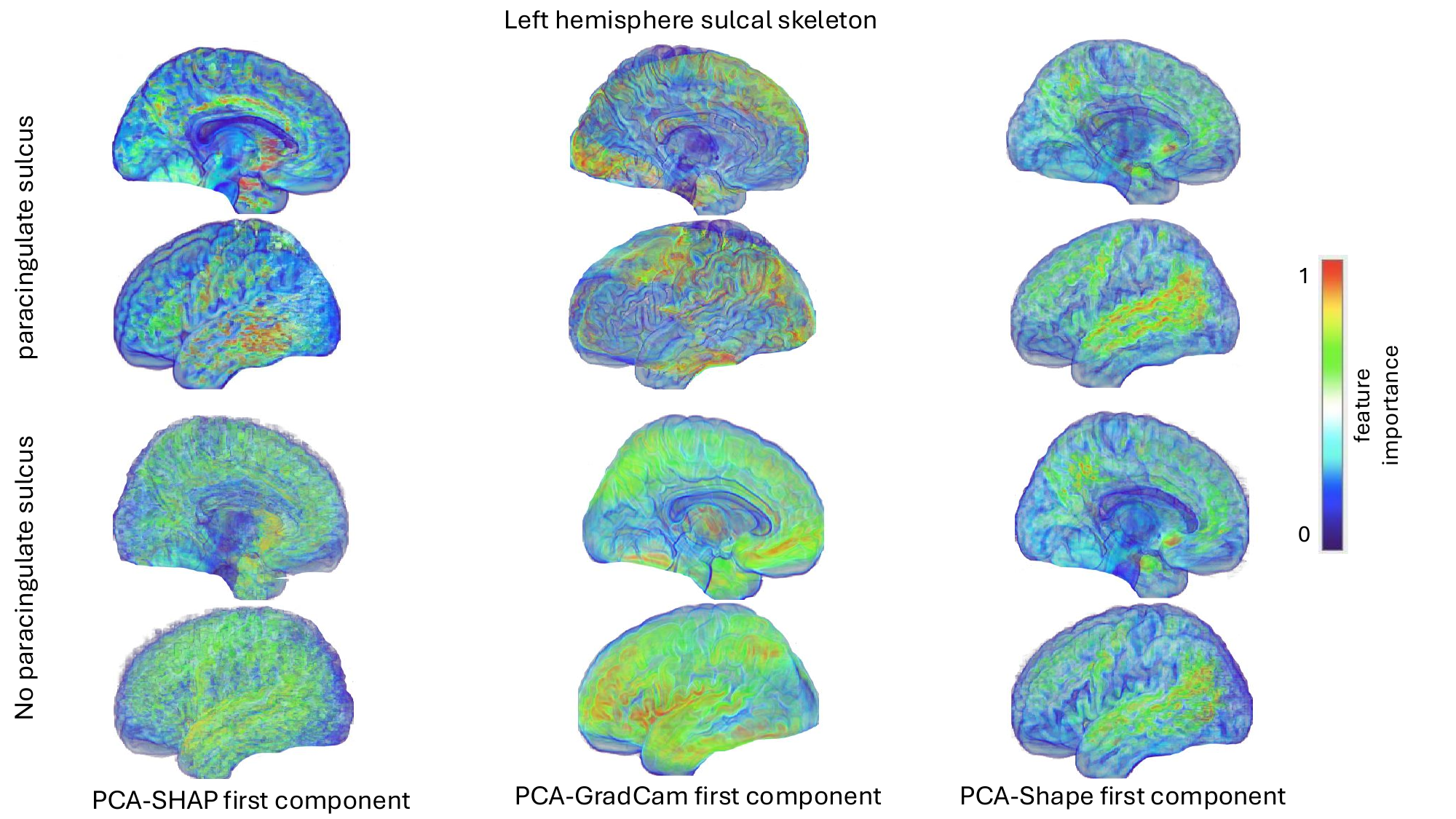}}
    \centerline{
\relax \textbf{b}
    \includegraphics[trim={0.0cm 0.0cm 0.0cm 0.0cm},clip,scale=.475]{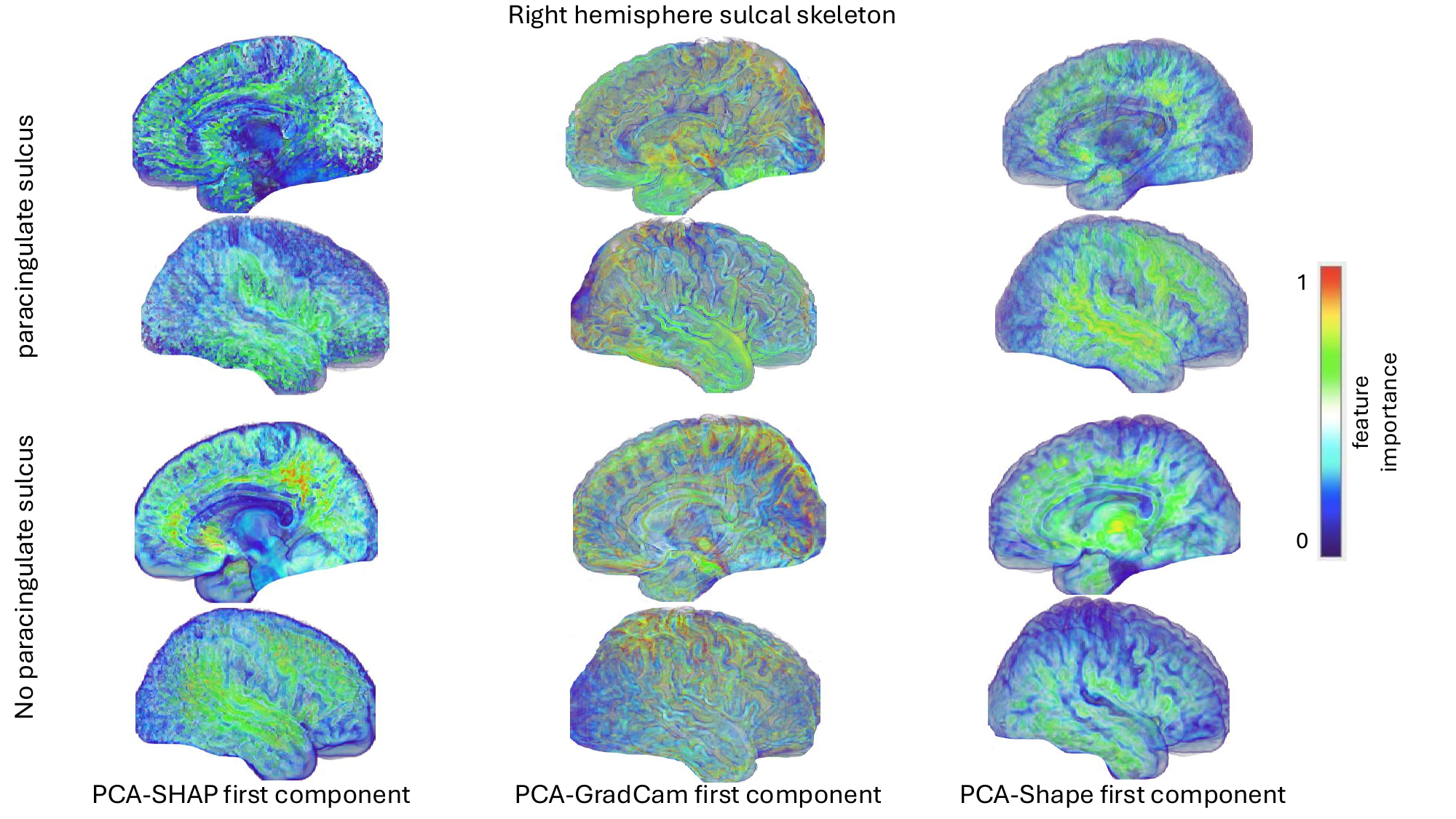}
    }

\caption{Simple-3D-MHL results on the left and right hemisphere of sulcal skeleton brain inputs.
\textbf{a-b,} show the explainability results for the noPCS class images of the first component among the six components of PCA for the total input modality (PCA-Shape), the total corresponding GradCam (PCA-GradCam), and the total corresponding SHAP results (PCA-SHAP). The feature's importance (pixel attribution) varies from 0 (blue color) to 1 (red color), with high importance being 1 for the PCA-GradCam and PCA-Shape results. The orientation of the results are based on the medial anatomical views. All the presented results are align and mapping in the ICBM 2009a Nonlinear Asymmetric atlas (\cite{atlas1,atlas2}).}
  \label{x23}
\end{figure}

 Fig. \ref{x23}.a,b displays the comprehensive explanations of PCA-GradCam, PCA-SHAP, and PCA-Shape for the simple-3D-MHL network when using the sulcal skeleton inputs. The neural network shows distributed attention but still emphasizes some key regions in both hemispheres. In the left hemisphere, for the PCS class, there is  focus on the superior temporal sulcus and its branches, the posterior sylvian fissure, and parts of the central and precentral sulci, with some attention on the cingulate sulcus and the medial frontal sulcus (containing the PCS) (Fig. \ref{x23}a.). Conversely, in the right hemisphere for the same class, emphasis is on the superior and inferior temporal sulci, cingulate sulcus, medial frontal sulcus (containing the PCS), and the sub-parietal sulcus (Fig. \ref{x23}b.). For the noPCS class, the left hemisphere shows similar yet less specific attention than for the PCS class, with an additional focus on the sylvian fissure and insula, and the anterior cingulate sulcus (Fig. \ref{x23}a.). In the right hemisphere, additional focus is shifted to the anterior cingulate sulcus, the sub-parietal sulcus, and elements of the ventricles (Fig. \ref{x23}b.).
\par To validate the pattern indicating the presence or absence of the PCS, we assessed the variability across all six components of the PCA. Notably, we observed differences in the six PCA components between the explanations and the global feature importance (refer to Supplementary  Fig. 4). There were differences in the intensity and extent of regions highlighted between the sulcal skeleton and grey-white surface inputs, although the primary regions of focus remained consistent. We found that the faithfulness and complexity score of the PCA's first component for SHAP and GradCam methods performed poorly compared to the weighted average output of all six components as described in subsection 3.4 (total-SHAP, total-GradCam). Consequently, we used total-SHAP, total-GradCam, and total-Shape to compute the global explanation for the 3D framework. 

\subsection{Global explanations from the 3D-Framework and the pattern learning results using grey-white surface inputs}
\par For the simple-3D-MHL on grey-white surface inputs (see Fig. \ref{xait}a.c), in the right hemisphere the focus was on the frontal lobe, the insula, and parietal lobe in the PCS existence and, in the PCS absence (noPCS) on the temporal lobe, frontal lobe, cingulate gyrus and parietal lobe. In the left hemisphere, the PCS decision mostly relied on the cingulate gyrus, frontal lobe parietal lobe, and corpus collosum. The noPCS condition predominantly focused on the frontal lobe, and lateral temporal lobe and cingulate gyrus (see Fig. \ref{xait}a.c).
\par For the simple-3D-CNN on grey-white surface inputs (refer to Supplementary material  Fig. 6 a.) in the right hemisphere, highlighted the lateral inferior frontal lobe, and inferior temporal lobe in the PCS condition and the thalamus, cingulate gyrus, lateral anterior occipital lobe, and posterior temporal lobe. In the noPCS condition, the focus was on the thalamus,  cingulate gyrus, the medial frontal lobe, and the posterior temporal lobe. In the left hemisphere, both conditions globally focused on the same regions: the lateral middle frontal lobe, and inferior and superior parietal lobe, and the thalamus and in the cingulate, frontal and medial parietal lobes, with a special focus on the  anterior cingulate gyrus. The PCS condition additionally focused on the lateral view of the posterior temporal lobe.
\par For both networks, in the right hemisphere the focus was primarily on the medial aspect of the brain (except for the PCS condition in the right hemisphere) with the main contributions in the frontal lobe and cingulate gyrus, suggesting a rather constrained explainability of the PCS presence. Conversely, in the left hemisphere, the focus was much more broadly distributed, with strong contributions stemming from the lateral cortex, suggesting that the developmental mechanisms leading to the presence of the PCS are related to the wider development of the brain (\cite{fornito}).
\begin{figure*}
\medskip
\centerline{
\relax \textbf{a}
    \includegraphics[trim={0.6cm 0.05cm 0.25cm 0.05cm},clip,scale=.5]{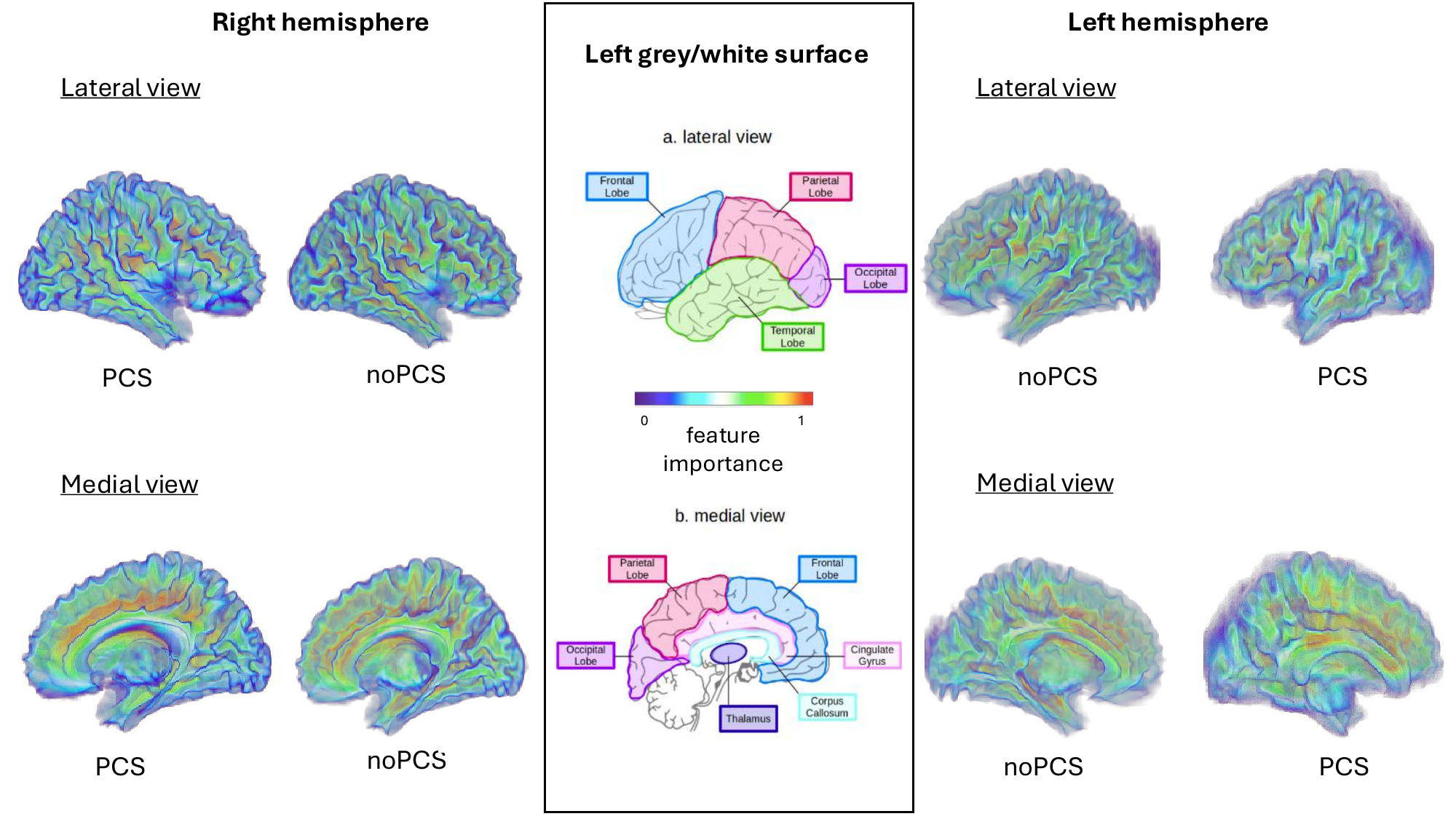}}
\centerline{
\relax \textbf{b}
    \includegraphics[trim={0.15cm 0.25cm 0.25cm 0.05cm},clip,scale=.5]{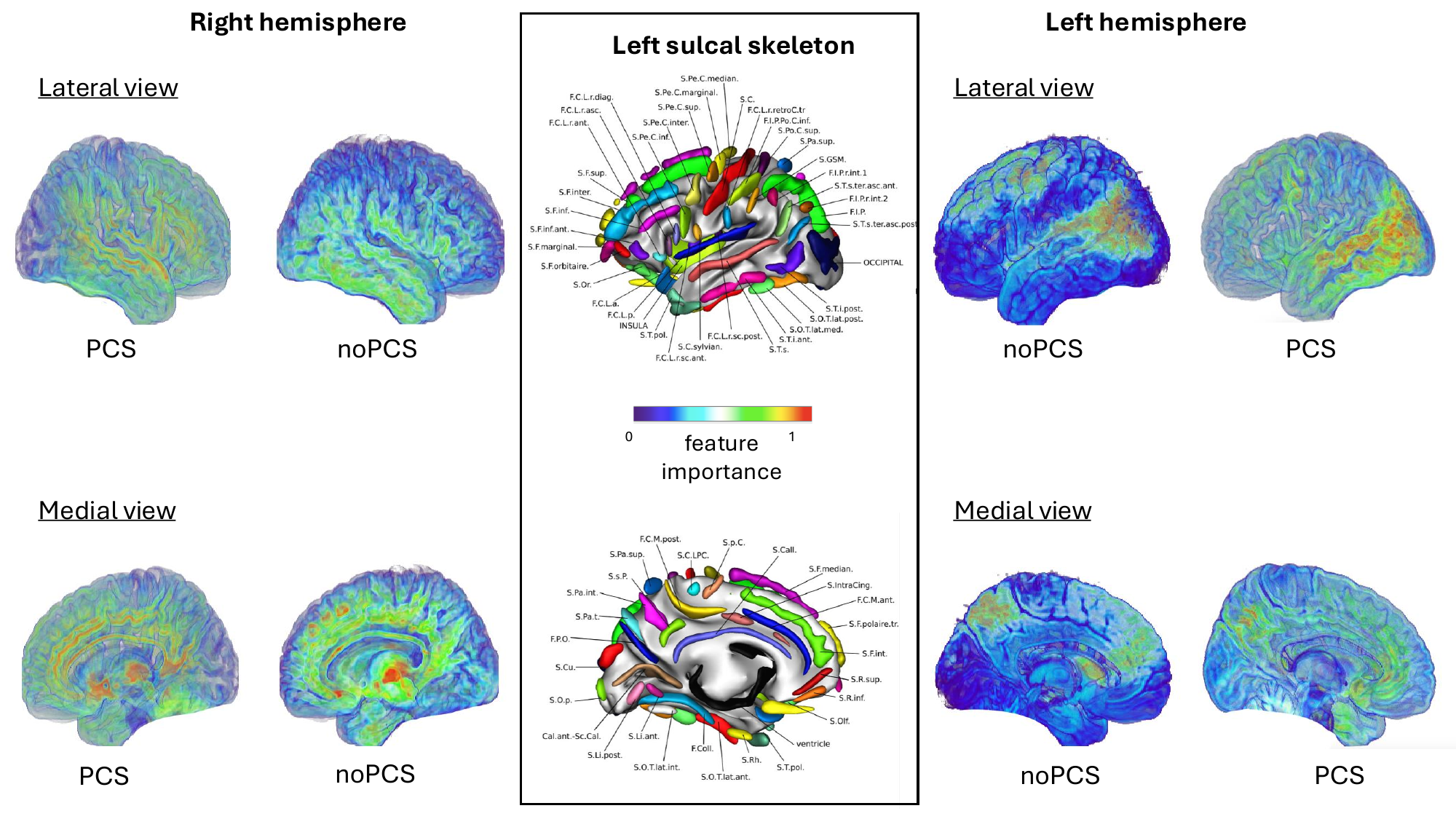}}
\centerline{
\relax \textbf{c}
    \includegraphics[trim={0.05cm 0.325cm 0.05cm 0.285cm},clip,scale=.665]{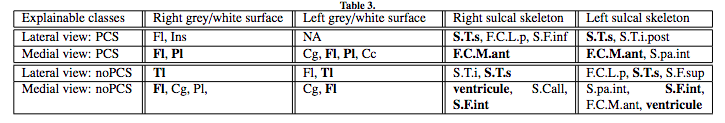}}    
\par\medskip
\end{figure*}
\begin{figure}
\caption{Explainable method's scores and the extraction of the total overlapping pattern learning explanation of the left and right hemispheres and different inputs with expert's observation.
\textbf{a} shows the total overlapping pattern learning results for the right and left hemisphere of the brain for grey-white surface images of the simple-3D-MHL network. \textbf{b} shows the total overlapping pattern learning results for the right and left hemispheres of the brain for sulcal skeleton inputs of the simple-3D-MHL network. 
All the presented results are align and mapping in the ICBM 2009a Nonlinear Asymmetric atlas (\cite{atlas1,atlas2}).
\textbf{c} shows the pattern learning results from the overlapping of the simple-3D-MHL deep learning networks on the total overlapping pattern learning results of the lateral and medial views based on experts' observation. For the grey-white surface input we used the acronyms, 
T: thalamus, H: hypothalamus, Fl: Frontal lobe, Ol: occipital lobe, Tl: temporal lobe, Pl: parietal lobe, Cc: corpus callosum, Cg: cingulate gyrus, NA: none. For the skeleton sulcal input we used the acronyms,
S.T.s: superior temporal sulcus,
S.T.i: inferior temporal sulcus,
F.C.M.ant: anterior cingulate sulcus,
S.pa.int:internal parietal sulcus,
F.C.L.p: sylvian fissure,
S.F.sup: superior frontal sulcus,
S.F.int: internal frontal sulcus,
S.F.inf: inferior frontal sulcus,
S.Call: callosal sulcus.}
  \label{xait}
\end{figure}
 \begin{table*}
\centering
\caption{Performance metrics of global explanation scores of faithfulness, and complexity for the global state of the art explanation methods; total-SHAP, total-GradCam and the propose 3D-Framework of the simple-3D-MHL network.}
\centering
\begin{tabular}{|p{2.75cm}||p{2.75cm}||p{2.75cm}|p{2.75cm}||p{2.85cm}||p{2.85cm}|}
\hline
 \multicolumn{6}{|c|}{Explainable evaluation score of used explainable methods} \\
\hline
 XAI metrics & XAI method &  Left sulcal skeleton & Left white/grey & Right sulcal skeleton & Right white/grey\\
 &  &  PCS / noPCS & PCS / noPCS & PCS / noPCS & PCS / noPCS\\
\hline
Faithfulness ($\uparrow$)  & total-SHAP &  0.105 / 0.200  &  0.092 / 0.113 & 0.103 / 0.102 & 0.073 / 0.088\\
             & total-GradCAM &  0.044 / 0.090 &  0.143 / 0.164 &  0.067 / 0.085 & 0.119 / 0.129\\
              & 3D-Framework  &  \textbf{0.223}/\textbf{0.274} &  \textbf{0.207}/\textbf{0.222} & \textbf{0.188}/\textbf{0.195} & \textbf{0.192}/\textbf{0.214}\\
              \hline 
Complexity ($\downarrow$) & total-SHAP  &  14.593 / 14.586 &  14.572 / 14.544 & \textbf{14.571}/\textbf{14.572} & \textbf{14.583}/\textbf{14.582} \\
             & total-GradCAM &  \textbf{14.473}/\textbf{14.555} &  \textbf{14.373}/\textbf{14.403} & 14.586 / 14.582 & 14.596 / 14.547\\
             & 3D-Framework &  14.584 / 14.563  &  14.582 / 14.587 & 14.579 / 14.573 &  14.587 / 14.587\\ 
\hline
\end{tabular}
\label{t3112}
\end{table*}
\subsection{Global explanation from the 3D-Framework and the pattern learning results using sulcal skeleton inputs}
\par A consistent method to overlay the outputs of the sulcal skeleton, similar to the process described for the grey-white surface outputs, was applied. For the simple-3D-MHL on sulcal skeleton inputs (Fig. \ref{xait}b,c), in the right hemisphere, the presence of PCS focuses on the lateral superior temporal sulcus, sylvian fissure, inferior frontal sulcus, and the anterior cingulate sulcus. The absence of PCS (noPCS; Fig. \ref{xait}c) focuses on the inferior temporal sulcus, superior temporal sulcus, ventricule, callosal sulcus, and inferior frontal sulcus. In the left hemisphere, the PCS and noPCS conditions had very different highlights. In the PCS condition, the main focuses were on the lateral posterior part of the lateral superior temporal sulcus, posterior 
inferior temporal sulcus, sylvian fissure, and the inferior parietal sulcus. In the noPCS condition, the main contributions were in the sylvian fissure, lateral superior temporal sulcus, superior frontal sulcus, internal parietal sulcus, anterior cingulate sulcus, and ventricle.
\par For the simple-3D-CNN on sulcal skeleton inputs, aimed at the accurate detection of PCS within the sulcal hemisphere inputs, the pivotal sub-regions encompassed the superior temporal sulcus, inferior precentral sulcus, sylvian fissure and sub-parietal sulcus (see Supplementary Material  Fig. 6b). Conversely, when PCS was absent, the critical sub-regions within the right hemisphere sulcal skeleton inputs encompassed the ventricule, superior temporal sulcus, internal parietal sulcus and rostral sulcus.  Transitioning to the left hemisphere, the sulcal skeleton inputs underscore the significance of the superior temporal sulcus the, sylvian fissure, and the internal parietal sulcus. When PCS was not present, the important left hemisphere sulcal skeleton inputs comprised the superior temporal sulcus, ventricle, inferior precentral sulcus, internal parietal sulcus (see Supplementary Material  Fig. 6b).
\par We thereafter identified the common regions between the two networks' outputs. The overlap results of the two networks, simple-3D-CNN and simple-3D-MHL, for the presence and absence of PCS (noPCS) reveal several common regions of interest in both the right and left hemispheres. For the presence of PCS in the right hemisphere, both networks highlight the lateral superior temporal sulcus and sylvian fissure. In the left hemisphere under the PCS condition, both models emphasize the superior temporal sulcus, the sylvian fissure, and the internal parietal sulcus. For the absence of PCS (noPCS) in the left and right hemisphere, the networks overlap in highlighting the ventricle, part of superior temporal sulcus and  the internal parietal sulcus.
\begin{table*}[h!]
\centering
\caption{Performance metrics assessing the global explanation scores of faithfulness and complexity were computed across various weight combinations assigned to the global XAI methods (total-SHAP, total-GradCam) and global feature extraction (total-Shape) within the proposed 3D-Framework for evaluating the global explanations of the simple-3D-MHL network. These combinations, denoted as 851, 815, 185, 158, 518, and 581, represent fixed-order representations of the assigned weights in the following sequence: total-Shape, total-SHAP, and total-GradCam explanations, with weight values of 0.85 represented as '8', 0.5 as '5', and 0.1 as '1'. For instance, 3D-Framework-851 refers to the proposed 3D framework with weight values of 0.85 in total-Shape, 0.5 in total-SHAP, and 0.1 in total-GradCam. The results shows the Right and Left white-grey (a) and sulcal skeleton (b) images.\\}
\centering
\begin{tabular}{|p{2.5cm}||p{3.5cm}||p{3.5cm}|p{3.5cm}|}
\hline
 \multicolumn{4}{|c|}{(a) Ablation study of the proposed 3D-Framework} \\
\hline
 XAI metrics & XAI method & Left white/grey  & Right white/grey\\
 &  &  PCS / noPCS & PCS / noPCS \\
\hline
Faithfulness  & 3D-Framework-851  &  0.166 / 0.197 & 0.172 / 0.182\\
              & 3D-Framework-815  & \textbf{0.207} / \textbf{0.222} & \textbf{0.192} / \textbf{0.214}\\
              & 3D-Framework-185  &  0.118 / 0.124  & 0.113 / 0.133\\
              & 3D-Framework-158  &  0.143 / 0.173 &  0.155 / 0.162\\
              & 3D-Framework-518  & 0.125  / 0.152  & 0.136 / 0.142\\
              & 3D-Framework-581   &  0.087 / 0.102  & 0.103 / 0.112\\
              \hline 
Complexity   & 3D-Framework-851 &   \textbf{14.582} / \textbf{14.582} &  14.595 / 14.592\\
              & 3D-Framework-815  &  14.582  / 14.587 & 14.587 / 14.587\\
              & 3D-Framework-185  & 14.585  / 14.582 & \textbf{14.584} / \textbf{14.587}\\
              & 3D-Framework-158  &  14.587 / 14.583 & 14.593 / 14.592\\
              & 3D-Framework-518  &  14.592 / 14.584 & 14.594 / 14.594\\
              & 3D-Framework-581  &  14.593 / 14.592 & 14.597 / 14.592\\   
\hline
\end{tabular}
\label{t311}
\end{table*}
\begin{table*}[h!]
\centering
\begin{tabular}{|p{2.5cm}||p{3.5cm}||p{3.5cm}|p{3.5cm}|}
\hline
 \multicolumn{4}{|c|}{(b) Ablation study of the proposed 3D-Framework} \\
\hline
 XAI metrics & XAI method &  Left sulcal skeleton & Right sulcal skeleton \\
 &  &  PCS / noPCS & PCS / noPCS \\
\hline
Faithfulness  & 3D-Framework-851&\textbf{0.223}/\textbf{0.274}& \textbf{0.188}/\textbf{0.195} \\
              & 3D-Framework-815  &  0.204 / 0.233  & 0.165 / 0.174\\
              & 3D-Framework-185  &  0.105 / 0.122  & 0.053 / 0.133 \\
              & 3D-Framework-158  &  0.074 / 0.095  & 0.122 / 0.076\\
              & 3D-Framework-518  &  0.115 / 0.106 & 0.144 / 0.134\\
              & 3D-Framework-581  &  0.126 / 0.138  & 0.135 / 0.142\\
              \hline 
Complexity   & 3D-Framework-851 &  14.584 / \textbf{14.563}  &  14.579 / \textbf{14.573}\\
              & 3D-Framework-815  &  14.594 / 14.595  & 14.582 / 14.576 \\
              & 3D-Framework-185  &  14.595 / 14.584  & \textbf{14.572} / 14.595 \\
              & 3D-Framework-158  &  14.592 / 14.583 & 14.573 / 14.594 \\
              & 3D-Framework-518  &  \textbf{14.583} / 14.585 & 14.573 / 14.576 \\
              & 3D-Framework-581  &  14.592 / 14.586  & 14.573 / 14.587 \\   
\hline
\end{tabular}
\label{t311}
\end{table*}

\subsection{Ablation study of our 3D explainability framework}
\par An ablation study in the simple-3D-MHL model was conducted to evaluate various combinations of global explanations (total-GradCam and total-SHAP) and global feature importance (total-Shape) (see Table \ref{t311}). The best results were achieved by assigning the highest weight (0.85) to total-Shape. This aligns with the idea that the feature importance of the inputs plays a crucial role in the explanations. For both hemispheres in the sulcal skeleton, a weight of 0.5 given to total-SHAP produced the highest faithfulness scores. For the grey-white surface inputs, total-GradCam with the same weight yielded the best results (a weight of 0.5). There were no significant differences observed in the complexity scores among the different combinations. The same patterns were observed for the simple-3D-CNN (Supplementary material; Table 2). 
\subsection{Evaluation of the Global explanation from the 3D-Framework and the XAI methods and the pattern learning results}
\par To evaluate whether the global explanation from the 3D-Framework was superior to those provided by SHAP or GradCam, we scored the explanations with respect to faithfulness and complexity (see Table \ref{t3112}). For the simple-3D-MHL network, our proposed 3D framework outperformed total-GradCam and total-SHAP in terms of faithfulness score in the left hemisphere with values exceeding 0.21 compared to scores of less than 0.16 for total-GradCam, and less than 0.11 for total-SHAP. In the right hemisphere, our proposed 3D framework again outperformed total-GradCam and total-SHAP, achieving faithfulness scores over 0.18 compared to scores of less than 0.13 for total-GradCam, and less than 0.10 for total-SHAP. The 3D framework achieved the second-best result in complexity scores with total-GradCam having the lowest score in the left hemisphere and total-SHAP in the right hemisphere.
\par Up to this point, we have mainly explored the explanation results visually. However, we aimed to automate the process to identify the most significant subregions of interest based on the hypothesis (the classification task). To this end, we applied an affine registration to the total overlapping results of sulcal skeleton inputs from each hemisphere onto a probabilistic atlas of sulci (\cite{perrot_cortical_2011}). For this task, we explored the sulcal skeleton output of the simple-3D-MHL network as it slightly outperformed the simple-3D-CNN (see Supplementary Material Table 2 ii.) in the classification task and delivered better global explanations, faithfulness, and complexity scores than the simple-3D-CNN. Additionally, it follows patterns based on evidence from the literature (\cite{fedeli, harper_structural_2024}, \cite{lopez-persem_human_2019}). 
\par
\begin{figure}
      \medskip     
\centerline{
    \relax \textbf{a}
    \includegraphics[trim={0.05cm 0.0cm 0.05cm 0.0cm},clip,scale=.545]{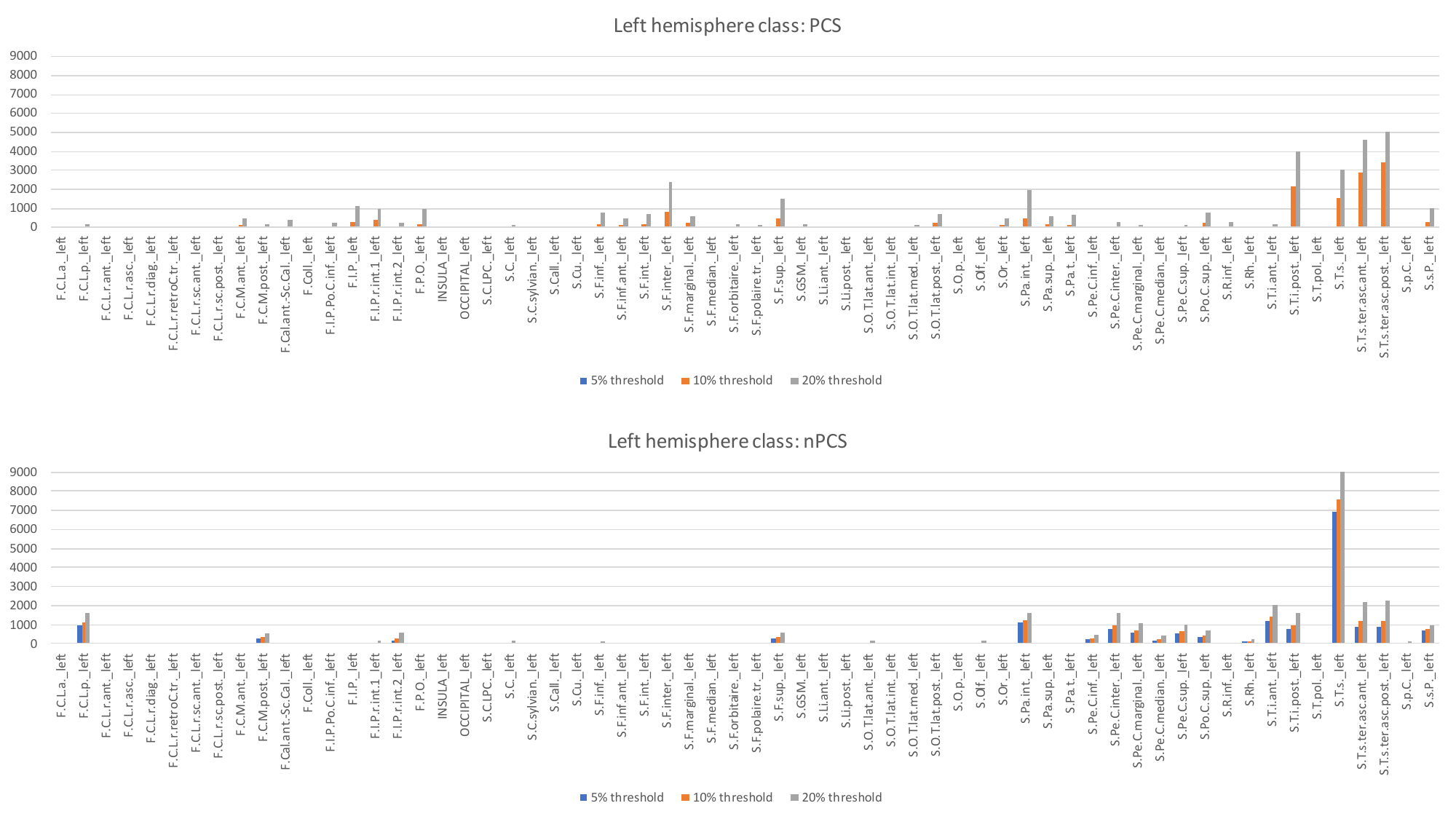}
}
\centerline{
\relax \textbf{b}
    \includegraphics[trim={0.00cm 0.0cm 0.0cm 0.0cm},clip,scale=0.35]{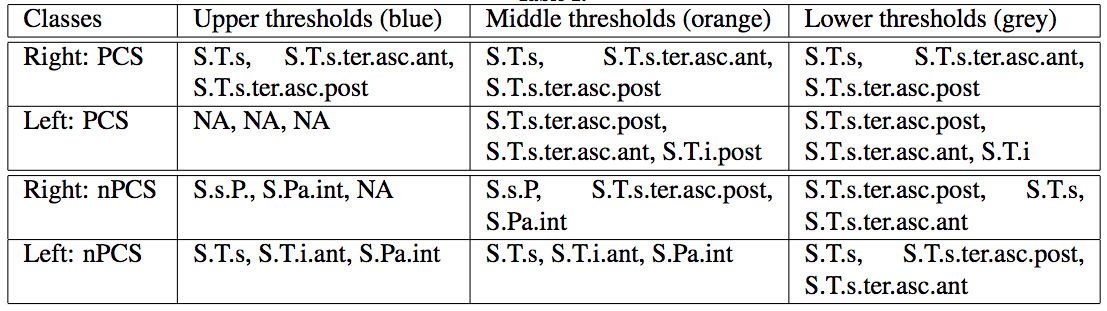}}
\par\medskip
\caption{A representation depicting the presence and absence of the paracingulate sulcus on the left hemisphere sulcal skeleton input
\textbf{a,} is the histogram of the number of voxels per sulcus based on the probabilistic mapping of sulci for both conditions (PCS and noPCS) and both hemispheres, using sulcal skeleton brain input images with the simple-3D-MHL network. The voxels are extracted after thresholding for highest explainability values, with thresholds of the highest 5, 10, and 20\% intensity on the left hemisphere (high threshold: blue, medium threshold: orange, low threshold: grey). \textbf{b,} the total overlapping pattern learning results of the three most significant sulcal sub-region for the three different level of intensity thresholding. The acronyms for all sulci are defined in Supplementary  Fig. 5., and NA: undefined.}
  \label{x2s}
\end{figure}

 Figure \ref{x2s} presents the distribution of the most relevant voxels for outcome decisions within sulcal probabilistic areas according to different thresholds. The decisions in the right hemisphere were highly focused on specific sulci, with up to three sulci contributing to the 20\% threshold, which we retained as the lower threshold (blue). Conversely, the left hemisphere predictions were based on broader considerations, with a number of sulci already contributing to the decision at the 5\% threshold, which we retained as the lower threshold for the left hemisphere (blue). 
\par For both conditions and both hemispheres, a specific focus was on the superior temporal sulcus and its posterior branches with a smaller but consistent contribution of the internal parietal and sub-parietal sulci. The noPCS condition on the left hemisphere additionally focused on the precentral sulcus and the Sylvian fissure. Interestingly, no specific focus was oriented towards the internal frontal sulcus (S.F.int), the probabilistic region in which the PCS is located when present.
\section{Discussion}
\par Explainability is essential in medical imaging. Healthcare applications of AI need to able to explain their decision making to build trust and ensure that their predictions align with other symptoms and signs that affect health. Neuroimaging, the combination of brain images and computational methods, is a research application of medical imaging. Here, explainability for AI predictions support the assessment of the validity of results, but can also identify key contributors to decisions that in themselves reveal new patterns and directions for future investigation.  
\par Our prior study (\cite{perspective}) categorized the need for AI explanations into self-explainable, semi-explainable, non-explainable applications, and new-pattern discovery, based on the variability of expert opinions, the stability of the evaluation protocol and the representation dimensionality of the application. We applied the proposed guidance in a binary classification task related to a sub-region of the medial surface of the brain where secondary sulci are highly reproducible related with symptoms of psychosis, specifically hallucinations. This application was of the new-pattern discovery class. The output of the explainability indicated a wide distribution of brain regions on which the predictions depend suggesting covariant development of these regions during the perinatal period (\cite{de_vareilles}).
\par Automatic classification of psychotic and control patients based on structural MRI is a challenging task (\cite{basic}). In most cases, an acceptable detection rate is around 80.0\% in clinical applications. However, in highly heterogeneous and variable cohorts, a lesser performance can be acceptable; approximately 60.0-70.0\% acceptance accuracy; \cite{thr1, th2}. The TOP-OSLO cohort was particularly difficult for classification tasks of psychotic and control patients using structural MRI, with an accuracy of less than 60.0\% (\cite{basic}). The variability of the paracingulate region in the TOP-OSLO cohort and the heterogeneity of the dataset create a highly challenging context for the automatic binary classification task of PCS presence. 
In this study, an accuracy of more than 70.0\% in the left hemisphere and more than 60.0\% in the right hemisphere was achieved, delivering an acceptable automated 3D deep learning network (\cite{thr1, th2}) to apply global explainability methods for new-pattern discovery (\cite{perspective}).
\par For the binary classification task of PCS presence, we developed two different 3D deep learning networks: a simple 3D convolutional neural network and a two-headed attention layer network. These networks utilized 3D brain inputs derived from preprocessed structural MRI scans, which included grey-white surface boundaries and sulcal skeletons from both hemispheres of a well-annotated cohort of 596 subjects. The performance of all networks was higher in the left hemisphere than in the right hemisphere. This discrepancy in performance was expected as the PCS is more prominent in the left hemisphere, including in psychopathological situations such as schizophrenia (\cite{garrison}). Moreover, the left PCS has more associations with regional cortical thickness and sulcal depth than the right PCS, implying greater covariability of anatomical features in our input modalities with the presence of the PCS in the left hemisphere (\cite{fornito}).
\par We developed an innovative XAI 3D-Framework to address the need for accurate, low-complexity global explanations in neuroimaging, where traditional 2D methods fall short in capturing the intricacies of 3D representational spaces. Designed for the binary classification of PCS presence, our framework provides robust, faithful global explanations that outperform GradCam and SHAP in faithfulness. Key novelties include the integration of statistical features (Shape) with reduced dimensionality information, ensuring explanations reflect both model learning and cohort-specific variability, and the combined use of GradCam and SHAP to reduce inter-method variability and enhance reliability. This multi-method framework sets a new standard for explainable AI in neuroimaging, offering actionable insights for complex tasks like cortical morphology analysis. Our global explanations surpassed those produced by GradCam and SHAP in terms of faithfulness, providing a reliable interpretation of the deep networks for this classification task.
\par The overall explainability outputs cover wide regions of the brain, but we can notice some repetitive patterns through the different pipelines and modalities. In particular, there is a repeat of the cingulate region, the posterior temporal region, and both the medial and inferior frontal cortices. This may reflect some neurodevelopmental intertwining of the macroscopical development of the PCS and these regions. Regarding the cingulate and medial frontal regions, this intertwining is self-explanatory as the PCS is located in the medial frontal region, directly adjacent to the cingulate region, and as such the developmental events leading to the formation of a PCS are very likely to affect these regions. The two other notable regions are the inferior frontal region and the posterior temporal region. In the fetus, the sulci matching these regions (namely the inferior frontal sulcus and the posterior superior temporal sulcus) have both been reported to start appearing at 26 weeks of gestational age (w GA), while the cingulate sulcus appears earlier (around 23w GA) and the “secondary cingular sulci”, which encompass the PCS, appear later, at 31w GA (\cite{garel_fetal_2001}). This may point towards a time-window which is decisive to the development of the PCS, prior to its actual apparition. 
\par In terms of functional interpretation, it is interesting to notice a striking similarity between the regions on which the AI mostly focuses and the regions which have been reported to show the most functional connectivity with the paracingulate region (\cite{fedeli, harper_structural_2024}), and the anterior cingulate region (\cite{lopez-persem_human_2019}). These studies report relevant functional connectivity between the medial frontal lobe (including the PCS) and the temporal region (including a focus on the posterior superior temporal region), the inferior frontal region, and the medial parietal region, which are all regions showing particular interest in the present work. Both \cite{fedeli} and \cite{harper_structural_2024} investigate the relation between the presence of a PCS and the related functional connectivity in these regions, and obtain more focused results (respectively in the cerebellum and superior anterior temporal region, or in the medial frontal region), but the important functional relationships between these regions support the relevance of the regions highlighted by our results.  
\par The effectiveness of extracting generalized patterns using our proposed framework underscores the importance of incorporating data from multiple cohorts. To this end, we plan to apply the framework to additional cohorts, such as BeneMin (\cite{ben}) and Biobank (\cite{bio}), to identify shared patterns in the classification of PCS presence and absence. Combining XAI techniques with dimensionality reduction methods may further reveal overlapping aspects of the data. Advanced approaches, such as t-SNE for non-linear dimensionality reduction, could also provide deeper insights into these relationships. Additionally, we aim to extend the framework’s application to other neurological conditions and classification tasks, including schizophrenia and bipolar disorder, by leveraging external datasets and improving interpretability techniques.
\par This study establishes a foundation for systematic exploration of sulcal variability through deep learning, with the potential to advance our understanding of cognitive and functional variability as well as pathological changes.
\subsection{Conclusion}
In this study, we classified the presence or absence of the paracingulate sulcus (PCS) in a diverse cohort of 596 structural MRIs using 3D deep convolutional neural networks and attention mechanisms. To address the lack of robust global explainability methods for 3D neuroimaging data, we developed an innovative XAI 3D-Framework. This framework provides accurate and low-complexity global explanations for PCS detection by integrating statistical features (Shape) with XAI methods (GradCam and SHAP) alongside reduced dimensionality information, ensuring that the explanations capture both model learning and cohort-specific variability. Furthermore, the combined application of GradCam and SHAP mitigates inter-method variability, thereby enhancing the reliability and robustness of the explanations.
\par Our framework outperformed established methods like GradCam and SHAP in faithfulness, enabling the robust identification of sub-regions critical for decision-making through a fusion of global explanations and statistical features. Key patterns identified include a focus on the posterior temporal and internal parietal regions on the sulcal skeleton, and on the cingulate region and thalamus when analyzing the grey-white surface. These findings indicate potential co-variation between these structures, likely underpinned by shared genetic or developmental mechanisms. Such insights hold significant implications for both neurodevelopmental and pathological research, providing a foundational framework for guiding future investigative trajectories.
\par Our work advances both deep learning and neuroscience by enabling automated, unbiased annotations and delivering unprecedented insights into sulcal variability and its functional or pathological relevance. The XAI 3D-Framework sets the stage for broader applications in medical imaging and other complex computer vision tasks, providing a foundation for comprehensive exploration of neuroanatomy and developmental mechanisms.

\section*{Data and Code Availability}

\subsection*{Data availability} 
This study used the dataset of the TOP-OSLO cohort (\cite{morch-johnsen}) which can be obtained from University of Oslo upon request, subject to a data transfer agreement.

\subsection*{Code availability} 
The code developed in this study is written in the Python programming language using pytorch, Keras, tensorflow (Python) libraries. For training and testing of deep learning networks, we have used an NVIDIA cluster with $4$ GPUs and $64$~GB RAM memory. The code is publicly available in \url{https://github.com/ece7048/3Dsulci}. 

\section*{Author Contributions}

M.M, H.V., P.L., J.S. and G.M. conceived the study. M.M. wrote the code and conducted the experiments. I.A., LE.MJ., H.V., A.AM., and S.C.M. collected and pre-processed the data cohort. M.M, and H.V. analyzed the data and results. M.M. contributed to pulling deep learning and XAI methods and conducted chart reviews. M.M  contributed to the experimental design and validation protocol. G.M.,J.S.,P.L., H.V. and M.M. was in charge of overall direction and planning. All authors contributed to the interpretation of the results. M.M., H.V. and P.L. visualized the study and extracted figures, and M.M and H.V. drafted the manuscript, which was reviewed, revised and approved by all authors.

\section*{Funding}
This study was supported by funding from the Medical Research Council, grant number: MR/W020025/1.

\section*{Declaration of Competing Interests}

GKM consults for ieso digital health. All other authors declare that they have no competing interests. 

\section*{Acknowledgements}

All research at the Department of Psychiatry in the University of Cambridge is supported by the NIHR Cambridge Biomedical Research Centre (NIHR203312) and the NIHR Applied Research Collaboration East of England. The views expressed are those of the author(s) and not necessarily those of the NIHR or the Department of Health and Social Care. This study was supported by funding from the Medical Research Council, grant number: MR/W020025/1. We acknowledge the use of the facilities of the Research Computing Services (RCS) of University of Cambridge, UK. GKM consults for ieso digital health. All other authors declare that they have no competing interests. 

\section*{Supplementary Material}
The manuscript has Supplementary Material.

\bibliographystyle{unsrtnat}
\bibliography{ref}  

\end{document}